\documentclass[12pt]{msml2021} 

\usepackage{amsmath,amsfonts,bm}

\def\eqref#1{equation~\ref{#1}}

\def\1{\bm{1}}

\DeclareMathAlphabet{\mathsfit}{\encodingdefault}{\sfdefault}{m}{sl}
\SetMathAlphabet{\mathsfit}{bold}{\encodingdefault}{\sfdefault}{bx}{n}

\newcommand{\R}{\mathbb{R}}

\usepackage[utf8]{inputenc} 
\usepackage[T1]{fontenc}    
\usepackage{hyperref}       
\usepackage{url}            
\usepackage{booktabs}       
\usepackage{amsfonts}       
\usepackage{nicefrac}       
\usepackage{microtype}      
\usepackage{bm}
\usepackage{wrapfig}
\usepackage[nameinlink]{cleveref}
\usepackage{hyperref}
\usepackage{xr}
\usepackage{color}
\usepackage{xcolor}
\usepackage[normalem]{ulem}
\usepackage{enumitem}
\usepackage[flushleft]{threeparttable}
\usepackage{rotating}
\usepackage{caption}

\usepackage{array}
\usepackage{soul}
\newcolumntype{P}[1]{>{\centering\arraybackslash}p{#1}}
\newcolumntype{M}[1]{>{\centering\arraybackslash}m{#1}}

\newcommand{\mask}{\bm{S}}
\newcommand{\M}{\bm{M}}
\newcommand{\Mhat}{\bm{X}}
\newcommand{\Mlatent}{\bm{Z}}

\newcommand{\bb}[1]{\bm{#1}}
\newcommand{\Rphi}{\bb{P}}
\newcommand{\Rpsi}{\bb{Q}}
\newcommand{\Phibold}{\mathbf{\Phi}}
\newcommand{\Psibold}{\mathbf{\Psi}}
\newcommand{\Phiprime}{{\mathbf{\Phi}^\everymodeprime}}
\newcommand{\Psiprime}{{\mathbf{\Psi}^\everymodeprime}}
\newcommand{\C}{\bb{C}}

\newcommand{\trace}{\mathrm{tr}}
\newcommand{\Lrows}{\bb{L}_{\mathrm{r}}}
\newcommand{\Lcols}{\bb{L}_{\mathrm{c}}}
\newcommand{\Lambdarows}{\bb{\Lambda}_{\mathrm{r}}}
\newcommand{\Lambdacols}{\bb{\Lambda}_{\mathrm{c}}}
\newcommand{\Grows}{\mathcal{G}_{\mathrm{r}}}
\newcommand{\Gcols}{\mathcal{G}_{\mathrm{c}}}
\newcommand{\Gprod}{\mathcal{G}}
\newcommand{\Ga}{\mathcal{G}_1}
\newcommand{\Gb}{\mathcal{G}_2}
\newcommand{\G}{\mathcal{G}}
\newcommand{\Gprime}{\mathcal{G}^\everymodeprime}
\newcommand{\Frows}{\bb{F}_{p}}
\newcommand{\Fcols}{\bb{G}_{q}}
\newcommand{\Ez}{E_{\mathrm{data}}}
\newcommand{\Ezoomout}{E_{\mathrm{z}}}

\newcommand{\PiM}{\bm{\Pi}}

\newcommand{\loss}{\ell}
\newcommand{\Edir}{{E}_{\mathrm{dir}}}

\newcommand{\Ediagrows}{{E}^{r}_{\mathrm{diag}}}
\newcommand{\Ediagcols}{{E}^{c}_{\mathrm{diag}}}

\newcommand{\dirweight}{\mu}
\newcommand{\diagweight}{\rho}

\newcommand{\pmax}{p_{\mathrm{max}}}
\newcommand{\qmax}{q_{\mathrm{max}}}

\newcommand{\pskip}{p_{\mathrm{skip}}}
\newcommand{\qskip}{q_{\mathrm{skip}}}
\newcommand*{\everymodeprime}{\ensuremath{\prime}}
\newcommand{\A}{\bb{A}}
\newcommand{\B}{\bb{B}}

\newcommand{\squarespace}{\square\,}
\renewcommand{\eqref}[1]{(\ref{#1})}

\definecolor{mygreen}{rgb}{0.094, 0.6, 0.094}

\newcommand{\gr}[1]{\textcolor{mygreen}{#1}}
\newcommand{\red}[1]{\textcolor{red}{#1}}

\newcommand{\add}[1]{\textcolor{mygreen}{#1}}

\makeatletter
 \let\Ginclude@graphics\@org@Ginclude@graphics 
\makeatother

\title[Spectral Geometric Matrix Completion]{Spectral Geometric Matrix Completion}

\msmlauthor{%
 \Name{Amit Boyarski*} \Email{amitboy@cs.technion.ac.il}\\
 \addr Technion, Israel.
 \AND
 \Name{Sanketh Vedula*} \Email{sanketh@cs.technion.ac.il}\\
 \addr Technion, Israel.%
 \AND
 \Name{Alex Bronstein} \Email{bron@cs.technion.ac.il}\\
 \addr Technion, Israel.%
}

\begin{document}

\maketitle

\begin{abstract}
Deep Matrix Factorization (DMF) is an emerging approach to the problem of matrix completion. Recent works have established that gradient descent applied to a DMF model induces an implicit regularization on the rank of the recovered matrix.  In this work we interpret the DMF model through the lens of spectral geometry. This allows us to incorporate explicit regularization without breaking the DMF structure, thus enjoying the best of both worlds. In particular, we focus on matrix completion problems with underlying geometric or topological relations between the rows and/or columns. Such relations are prevalent in matrix completion problems that arise in many applications, such as recommender systems and drug-target interaction. Our contributions enable DMF models to exploit these relations, and make them competitive on real benchmarks, while  exhibiting one of the first successful applications of deep linear networks.
\end{abstract}

\begin{keywords}%
  matrix completion, deep matrix factorization, deep linear networks, graph signal processing, spectral geometry, recommendation system, drug-target interaction.
\end{keywords}

\section{Introduction}

Matrix completion deals with the recovery of missing values of a matrix from a subset of its entries,
\begin{equation}\label{eq:matrix_completion}
    \mathrm{Find}\;\;\Mhat\;\;\mathrm{s.t.}\;\; \Mhat\odot\mask = \M \odot \mask.
\end{equation}
Here $\Mhat$ stands for the unknown matrix, $\M\in\mathbb{R}^{m\times n}$ for the ground truth matrix, $\mask$ is a binary mask representing the input support, and $\odot$ denotes the Hadamard product. Since problem \eqref{eq:matrix_completion} is ill-posed, it is common to assume that $\M$ belongs to some low dimensional subspace. Under this assumption, the matrix completion problem can be cast via the least-squares variant,
\begin{equation}\label{eq:minRankLS}
    \min_{\Mhat}\; \mathrm{rank}\left(\Mhat\right)+\frac{\mu}{2}\left\|\left(\Mhat- \M\right) \odot \mask\right\|_F^2.
\end{equation}
Relaxing the intractable rank penalty to its convex envelope, namely the nuclear norm, leads to a convex problem whose solution coincides with \eqref{eq:minRankLS} under some technical conditions \citep{candes2009exact}. Another way to enforce low rank is by explicitly parametrizing $\Mhat$ in factorized form, $\Mhat = \bb{X}_1\bb{X}_2$.
The rank is upper-bounded by the minimal dimension of $\bb{X}_1,\bb{X}_2$.
Further developing this idea, $\Mhat$ can be parametrized as a product of several matrices $\Mhat = \prod_{i=1}^N \bb{X}_i$, a model we denote as \textit{deep matrix factorization} (DMF). Following the nomenclature used by \cite{arora2019implicit}, a factorization with $N> 2$ is called \textit{deep} while for $N\leq 2$ it is called \textit{shallow}. Nevertheless, to simplify notation, we shall simply call any such model as DMF while explicitly specifying $N$. \citet{gunasekar2017implicit,arora2019implicit} investigated the minimization of overparametrized DMF models using gradient descent, and came to the following conclusion (which we will formally state in \Cref{sec:Preliminaries}): whereas in some restrictive settings minimizing DMF using gradient descent is equivalent to nuclear norm minimization (i.e., convex relaxation of \eqref{eq:minRankLS}), in general these two models produce different results, with the former enforcing a stronger regularization on the rank of $\Mhat$. This regularization gets stronger as $N$ (the depth) increases. In light of these results, we shall henceforth refer by "DMF" to the aforementioned model coupled with the specific algorithm used for its minimization, namely, gradient descent. 

In this work we focus on matrix completion problems with underlying geometric or topological relations between the rows and/or columns. Such relations are prevalent in matrix completion problems that arise in applications such as recommender systems and drug-target interaction. A common way of representing these relations is in the form of a graph. For example, in the Netflix problem \citep{candes2009exact} the rows correspond to users, the columns correspond to movies, and the matrix elements represent the ratings given by the users to the movies. The goal is to recover the full matrix of ratings from an incomplete matrix. In this setting there might exist side information on the users and movies which can be used to construct two graphs representing relations between users and relations between movies, and the matrix $\Mhat$ can be viewed as a signal on the \textit{product} of these graphs. A useful prior on $\Mhat$ can be, for example, modeling it as a smooth or band-limited signal on this graph, encouraging similar movies to be assigned similar rating from similar users, and vice versa. 

This kind of geometric structure is generally overlooked by purely algebraic entities such as rank, and becomes invaluable in the data poor regime, where the theorems governing reconstruction guarantees (i.e., \citet{candes2009exact}) do not hold. Our work leverages the recent advances in DMF theory to marry the two concepts: a framework for matrix completion that is \textit{explicitly} motivated by geometric considerations, while \textit{implicitly} promoting low-rank via its DMF structure. 

\begin{figure*}[hb!]
    \centering
    \includegraphics[width=0.85\textwidth]{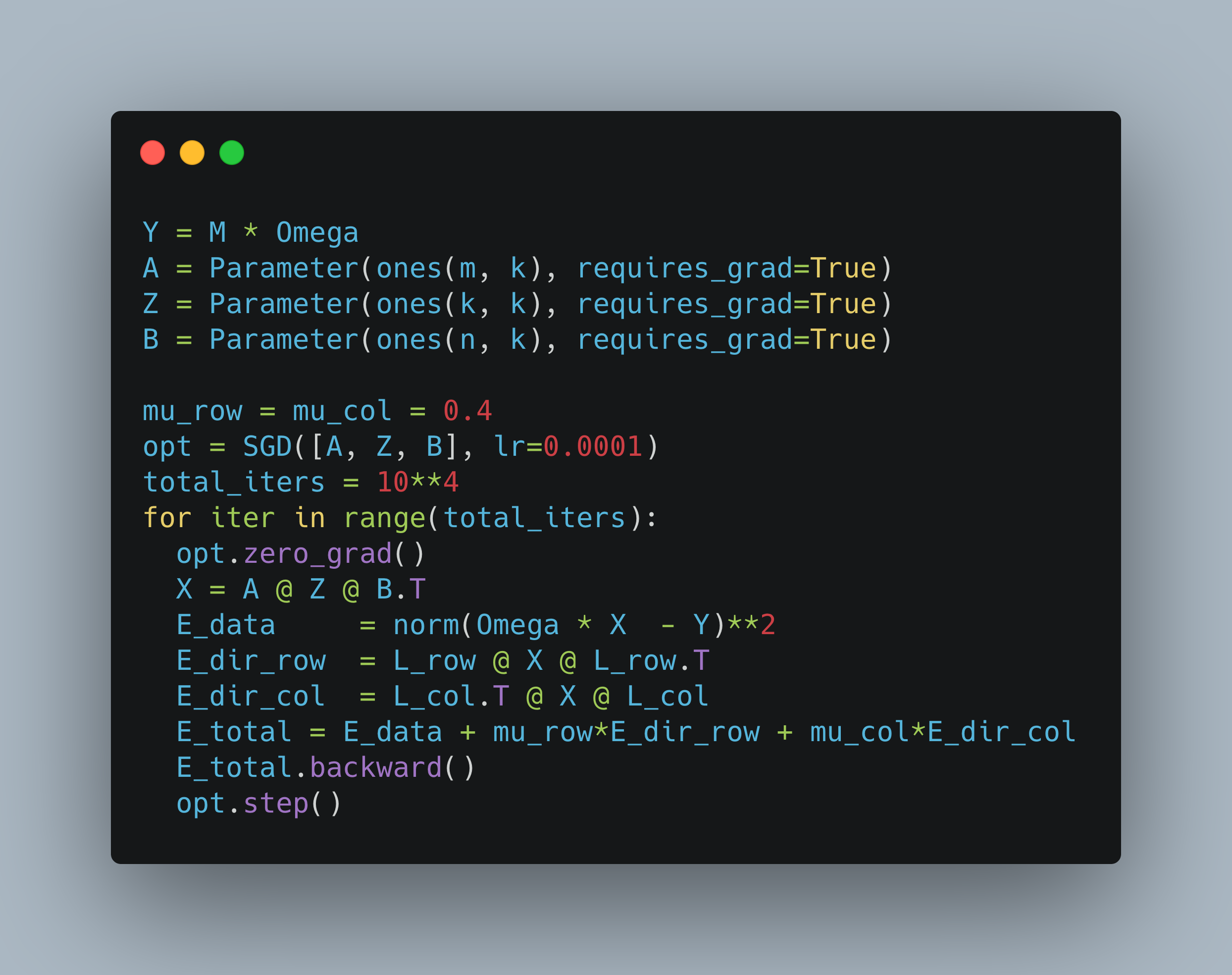}
    \caption{A few lines of code.\label{fig:code}}
\end{figure*}

\paragraph{Contributions.}
 Our contributions are as follows:
\begin{itemize}
    \item We propose geometrically inspired DMF models for matrix completion and study their dynamics. 

    
    \item We successfully apply those models to matrix completion problems in recommendation systems and drug-target interaction, outperforming various complicated methods with only a few lines of code (\Cref{fig:code}). This serves as an example to the power of deep linear networks, being one of their first successful applications to real problems.
    
    \item Our findings challenge the quality of the side information available in recommendation systems datasets, and the ability of contemporary methods to utilize it in a meaningful and efficient way.
    
\end{itemize}

\section{Preliminaries}
\label{sec:Preliminaries}

\paragraph{Spectral graph theory.}
Let $\G = (V,E,\bb{\Omega})$ be a (weighted) graph specified by its vertex set $V$ and edge set $E$, with its adjacency matrix denoted by $\bb{\Omega}$. Given a function $\bb{x}\in \mathbb{R}^{|V|}$ on the vertices, we define the following quadratic form (also known as \textit{Dirichlet energy}) measuring the variability of the function $\bb{x}$ on the graph,
\begin{equation}\label{eq:quadform}
    \bb{x}^\top \bb{L}\bb{x} = \sum_{(a,b)\in E}\omega_{a,b}\left(x(a)-x(b)\right)^2.
\end{equation}

The matrix $\bb{L}$ is called the \textit{(combinatorial) graph Laplacian}, and is given by $\bb{L} = \bb{D}-\bb{\Omega}$,
where $\bb{D} = \mathrm{diag}(\bb{\Omega}\bb{1})$ is the \textit{degree matrix}, with $\bb{1}$ denoting the vector of all ones. $\bb{L}$ is symmetric and positive semi-definite and therefore admits a spectral decomposition $\bb{L} = \Phibold\bb{\Lambda}\Phibold^\top$. Since $\bb{L}\bb{1}=\bb{0}$, $\lambda_1=0$ is always an eigenvalue of $\bb{L}$. The graph Laplacian is a discrete generalization of the continuous Laplace-Beltrami operator, and therefore has similar properties.
One can think of the eigenpairs $(\bb{\phi}_i,\lambda_i)$ as the graph analogues of "harmonic" and "frequency". 
Structural information about the graph is encoded in the spectrum of the Laplacian. For example, the number of connected components in the graph is given by the multiplicity of the zero eigenvalue, and the second eigenvalue (counting multiple eigenvalues separately) is a measure for the connectivity of the graph \citep{spielman2009spectral}.

A function $\bb{x}=\sum_{i=1}^{|V|}\alpha_i\bb{\phi}_i$ on the vertices of the graph whose coefficients $\alpha_i$ are small for large $i$, demonstrates a "smooth" behaviour on the graph in the sense that the function values on nearby nodes will be similar. A standard approach to promoting such smooth functions on graphs is by using the Dirichlet energy \eqref{eq:quadform} to regularize some loss term. For example, this approach gives rise to the popular bilateral and non-local means filters \citep{singer2009diffusion, gadde2013bilateral}. We call $\bb{x}$ a $k$-bandlimited signal on the graph $\G$ if $\alpha_i=0\;\forall\;i>k$.

\vspace{-0.0cm}
\paragraph{Product graphs and functional maps.}
Let $\Ga = \left(V_1,E_1,\bb{\Omega}_1\right)$, $\Gb = \left(V_2,E_2,\bb{\Omega}_2\right)$ be two graphs, with $\bb{L}_1= \Phibold\bb{\Lambda}_1\Phibold^\top$, $\bb{L}_2 = \Psibold\bb{\Lambda}_2\Psibold^\top$ being their corresponding graph Laplacians. The bases $\Phibold,\Psibold$ can be used to represent functions on these graphs. We define the Cartesian product of $\Ga$ and $\Gb$, denoted by $\Ga\squarespace \Gb$, as the graph with vertex set $V_1\times V_2$, on which two nodes $(u,u'),(v,v')$ are adjacent if either $u=v$ and $(u',v')\in E_2$ or $u'=v'$ and $(u,v)\in E_1$. The Laplacian of $\Ga\squarespace \Gb$ is given by the tensor sum of $\bb{L}_1$ and $\bb{L}_2$, 
\begin{equation}\label{eq:Dirichlet_prod_graph}
    \bb{L}_{\Ga\squarespace \Gb} =\bb{L}_1\oplus \bb{L}_2 =   \bb{L}_1\otimes \bb{I}+ \bb{I}\otimes \bb{L}_2,
\end{equation}
and its eigenvalues are given by the Cartesian sum of the eigenvalues of $\bb{L}_1,\bb{L}_2$, i.e., all combinations $\lambda_1+\lambda_2$ where $\lambda_1$ is an eigenvalue of $\bb{L}_1$ and $\lambda_2$ is an eigenvalue of $\bb{L}_2$.
Let $\Mhat$ be a function defined on $\Ga\squarespace \Gb$. Then it can be represented using the bases $\Phibold,\Psibold$ of the individual Laplacians, $\bb{C} = \Phibold^\top\Mhat\Psibold$. 
In the shape processing community, such $\bb{C}$ is called a \textit{functional map} \citep{ovsjanikov2012functional}, as it it used to map between the functional spaces of $\Ga$ and $\Gb$. For example, given two functions, $\bb{x} = \Phibold\bb{\alpha}$ on $\Ga$ and $\bb{y} = \Psibold\bb{\beta}$ on $\Gb$, one can use $\bb{C}$ to map between their representations $\bb{\alpha}$ and $\bb{\beta}$, i.e., $\bb{\alpha} = \Phibold^\top\bb{x}=\bb{C}\Psibold^\top\bb{y} = \bb{C}\bb{\beta}$. We shall henceforth interchangeably switch between the terms "signal on the product graph" and "functional map".

We will call a functional map \textit{smooth} if it maps close points on one graph to close points on the other. A simple way to construct a smooth map is via a linear combination of eigenvectors of $\bb{L}_{\Ga\squarespace \Gb}$ corresponding to small eigenvalues ("low frequencies"). Notice that while the singular vectors of $\bb{L}_{\Ga\squarespace \Gb}$ are outer products of the columns of $\Phibold$ and $\Psibold$, their ordering with respect to the eigenvalues of $\bb{L}_{\Ga\squarespace \Gb}$ might be different than their lexicographic order.
\paragraph{Implicit regularization of DMF.} 
Let $\Mhat\in \mathbb{R}^{m\times n}$ be a matrix parametrized as a product of $N$ matrices $\Mhat = \prod_{i=1}^N \bb{X}_i$ (which can be interpreted as $N$ linear layers of a neural network), and let $\loss(\Mhat)$ be an analytic loss function. We are interested in the following optimization problem,
\begin{equation}\label{eq:DMFmodel}
    \min_{\bb{X}_1,\ldots,\bb{X}_N}\loss\left(\prod_{i=1}^N \bb{X}_i\right).
\end{equation}
Without loss of generality, we will assume that $m<n$. \citet{arora2018optimization, arora2019implicit} analyzed the evolution of the singular values and singular vectors of $\Mhat$ throughout the gradient flow $\dot{\Mhat}(t) = -\nabla \loss(\Mhat(t))$, i.e., gradient descent with an infinitesimal step size, with \textit{balanced initialization},
\begin{equation}\label{eq:balanced}
    \Mhat_{i+1}(0)^\top \Mhat_{i+1}(0) = \Mhat_{i}(0) \Mhat_{i}(0)^\top,\;\;\forall i=1\ldots N.
\end{equation}
As a first step, we state that~$\Mhat(t)$ admits an analytic singular value decomposition.
\begin{lemma}(Lemma 1 in \citet{arora2019implicit}). \label{lemma:asvd}
The product matrix~$\Mhat(t)$ can be expressed as:
\begin{equation}
\Mhat(t) = \bb{U}(t) \bb{S}(t) \bb{V}^\top(t)
\text{\,,}
\label{eq:asvd}
\end{equation}
where:
$\bb{U}(t) \in \R^{m\times m}$, $\bb{S}(t) \in \R^{m\times m}$ and $\bb{V}(t) \in \R^{n\times m}$ are analytic functions of~$t$;
and for every~$t$, the matrices $\bb{U}(t)$~and~$\bb{V}(t)$ have orthonormal columns, while $\bb{S}(t)$ is diagonal (elements on its diagonal may be negative and may appear in any order).
\end{lemma}

The diagonal elements of~$\bb{S}(t)$, which we denote by $\sigma_1(t), \ldots, \sigma_{m}(t)$, are signed singular values of~$\Mhat(t)$;
the columns of $\bb{U}(t)$ and~$\bb{V}(t)$, denoted ${u}_1(t), \ldots, {u}_{m}(t)$ and ${v}_1(t), \ldots, {v}_{m}(t)$, are the corresponding left and right singular vectors (respectively). Using the above lemma, \citet{arora2019implicit} characterized the evolution of singular values as follows:

\begin{theorem}(Theorem 3 in \citet{arora2019implicit}). \label{thm:sing_vals_evolve}
The signed singular values of the product matrix~$\Mhat(t)$ evolve by:
\begin{equation}
\begin{split}
 &\dot{\sigma}_r(t) = -N \cdot \big( \sigma_r^2(t) \big)^{1-1/N} \cdot \langle \nabla\loss(\Mhat(t)),  \mathbf{u}_r(t) \mathbf{v}_r^\top(t) \rangle, \\
&r = 1, \ldots, m.
\label{eq:S_evolve}
\end{split}
\end{equation}
If the matrix factorization is non-degenerate, i.e., has depth~$N \geq 2$, the singular values need not be signed (we may assume $\sigma_r(t) \geq 0$ for all~$t$).
\end{theorem}
The above theorem implies that the evolution rates of the singular values are dependent on their magnitude exponentiated by $2-2/N$.
Ignoring the term $\langle \nabla\loss(\Mhat(t)),  \mathbf{u}_r(t) \mathbf{v}_r^\top(t) \rangle$, as $N$ increases,the evolution rate of the large singular values is enhanced while the evolution rate of the small ones is dampened. The increasing gap between the evolution rates of the large and small singular values induces an implicit regularization on the effective rank of $\Mhat(t)$. However, the evolution of the singular values also depends on the gradient of the loss function via the term $\langle \nabla\loss(\Mhat(t)), \mathbf{u}_r(t) \mathbf{v}_r^\top(t) \rangle$, making the choice of the loss consequential. While exact analysis of \eqref{eq:S_evolve} is hard in the absence of perfect characterization of this term, one can still leverage these dynamics by empirically exploring different loss functions.
\section{Spectral geometric matrix completion}
\label{sec:our_approach}
We assume that we are given a set of samples from the unknown matrix $\bb{M}\in\mathbb{R}^{m\times n}$, encoded as a binary mask $\bb{S}$, and two graphs $\Grows,\Gcols$, encoding relations between the rows and the columns, respectively. Denote the Laplacians of these graphs and their spectral decompositions by $\Lrows= \Phibold\Lambdarows\Phibold^\top$, $\Lcols = \Psibold\Lambdacols\Psibold^\top$.  
We denote the Cartesian product between $\Gcols$ and $\Grows$ by $\Gprod\equiv\mathcal{\Gcols}\squarespace\mathcal{\Grows}$, and will henceforth refer to it as our reference graph. Our approach relies on a minimization problem of the form
\begin{equation}\label{eq:initialOpt}
\begin{split}
    &\min_{\Mhat}\; \Ez(\Mhat)+\mu \Edir(\Mhat)\\
    &\mathrm{s.t.}\;\;\;\mathrm{rank}(\Mhat)\leq r,
\end{split}
\end{equation}
with $\Ez$ denoting a data term of the form
\begin{equation}
\Ez(\Mhat) = \left\|\left(\Mhat- \M\right) \odot \mask\right\|_F^2,
\end{equation}
 and $\Edir$ is the Dirichlet energy of $\Mhat$ on $\Gprod$, given by (see \eqref{eq:Dirichlet_prod_graph})\footnote{Note that it is possible to weigh the two terms differently, as we do in some of our experiments.}   
\begin{equation}\label{eq:dir_prod}
\Edir(\Mhat) = \trace\left(\Mhat^\top\Lrows\Mhat\right)+ \trace\left(\Mhat\Lcols\Mhat^\top\right).    
\end{equation}
To that end, we parametrize $\Mhat$ via a matrix product
$\Mhat = \A\Mlatent\B^\top$, and discard the rank constraint,
\begin{equation}\label{eq:DMFOpt}
\begin{split}
    \min_{\A,\Mlatent,\B}\; \Ez(\A\Mlatent\B^\top)+
    \mu \Edir\left(\A\Mlatent\B^\top\right).
\end{split}
\end{equation}
Since \eqref{eq:DMFOpt} is now a DMF model of the form \eqref{eq:DMFmodel},  according to \Cref{thm:sing_vals_evolve} the discarded rank constraint will be captured by the implicit regularization induced by gradient descent even if the factors are full size matrices. To emphasize this point, in most of our experiments we used $\bb{A}$ of size $m\times m$, $\bb{Z}$ of size $m \times n$ and $\bb{B}$ of size $n\times n$.

\begin{figure}
\centering
\includegraphics[width=0.5\textwidth]{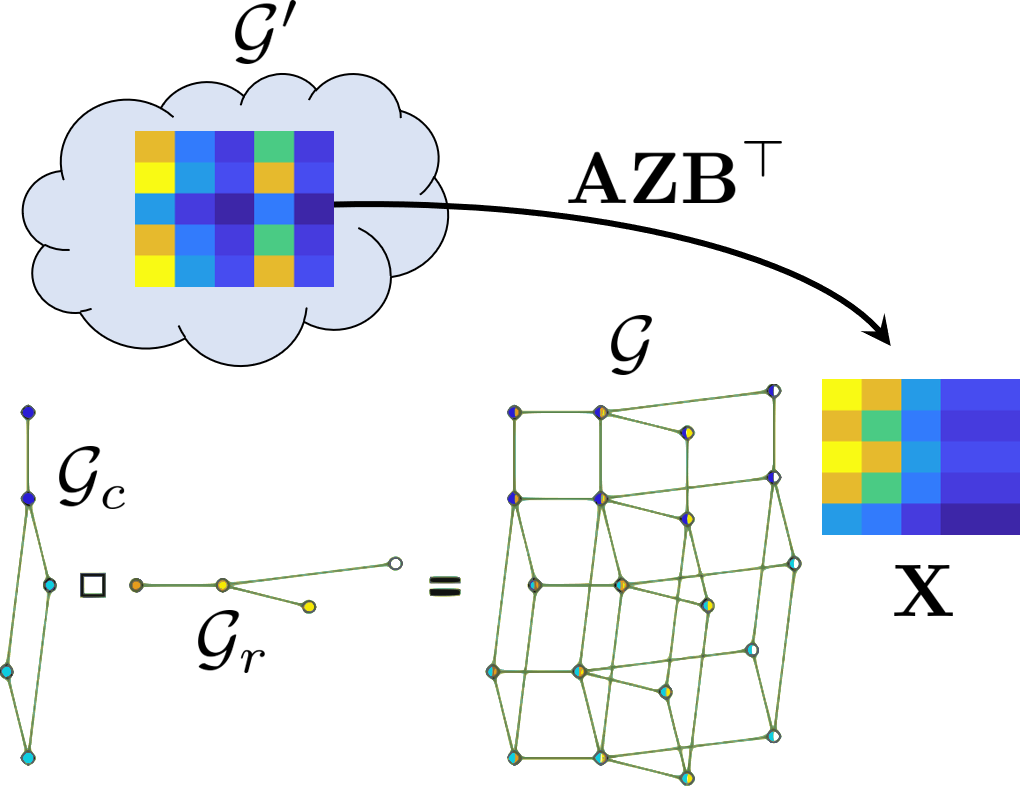}
\caption{\small An illustration of the geometric interpretation of \eqref{eq:DMFOpt}. A low-rank signal $\Mlatent$ that lives on a latent product graph $\Gprime$ is transported onto the reference product graph $\Gprod$. The transported signal $\A\Mlatent\B^\top$ will be smooth on the target graph due to the Dirichlet energy.}
\label{fig:prod_graph_transformation}
\end{figure}

To interpret this matrix factorization geometrically, we interpret $\Mlatent$ as a signal living on a latent product graph $\Gprime$. Via the linear transformation $\A\Mlatent\B^\top$ this signal is transported onto the reference graph $\Gprod$, where it is assumed to be both low-rank and smooth (see \Cref{fig:prod_graph_transformation}). Notice that the latent graph is used only for the purpose of illustrating the geometric interpretation, and there is no need to find it explicitly. Nevertheless, it is possible to promote particular properties of it via spectral constraints that can sometime improve the performance. We demonstrate these extensions in the sequel.

To give a concrete example, suppose $\Mhat$ is a band-limited signal on a 2D Euclidean grid $\Gprod$, then there is some low rank signal $\Mlatent$ that can be made smooth on $\Gprod$ via an appropriate ordering of its rows and columns\footnote{On a side-note, that is exactly the goal of the well known and closely related seriation problem \citep{recanati2018relaxations}.}, i.e., $\Mhat=\PiM_1\Mlatent\PiM_2^\top$. By smooth we mean that it has low Dirichlet energy \eqref{eq:quadform}, where $\bb{L}$ is the \add{discrete} 2D Euclidean Laplacian, i.e., $\trace\left(\bb{X}^\top\bb{L}\bb{X}\right) = \sum_{i,j}\left(x_{i+1,j}-x_{i,j}\right)^2+\left(x_{i,j+1}-x_{i,j}\right)^2$. 

For later reference let us rewrite \eqref{eq:DMFOpt} in the spectral domain. We will denote the Laplacians of the latent graph factors comprising $\Gprod$ by $\Lrows^\everymodeprime,\Lcols^\everymodeprime$ and their eigenbases by $\Phiprime,\Psiprime$. Using those eigenbases and the eigenbases of the reference Laplacians $\Lrows,\Lcols$, we can write,
\begin{align}
    &\Mlatent = \Phibold^\everymodeprime\C{\Psibold^\everymodeprime}^\top\label{eq:Zspect}\\
    &\A = \Phibold\Rphi{\Phibold^\everymodeprime}^\top\label{eq:Aspect}\\
    &\B = \Psibold\Rpsi{\Psibold^\everymodeprime}^\top\label{eq:Bspect}.
\end{align}
Under this reparametrization we get\
\begin{equation}
    \A\Mlatent\B^\top = \Phibold\Rphi\C\Rpsi^\top\Psibold^\top.
\end{equation}
With some abuse of notation, \eqref{eq:DMFOpt} becomes
\begin{equation}\label{eq:DMFOpt_spect}
\begin{split}
    \min_{\Rphi,\C,\Rpsi}\; \Ez(\Rphi\C\Rpsi^\top)+
    \mu \Edir\left(\Rphi\C\Rpsi^\top\right),
\end{split}
\end{equation}

with 
\begin{equation}\label{eq:dir_prod_spectral}
\Ez(\Rphi\C\Rpsi^\top) = \left\|\left(\Phibold\Rphi\C\Rpsi^\top\Psibold^\top- \M\right) \odot \mask\right\|_F^2,
\end{equation}

and
\begin{equation}
    \begin{split}
        \Edir(\Rphi\C\Rpsi^\top) &= \trace\left(\Rpsi\C^\top\Rphi^\top\Lambdarows\Rphi\C\Rpsi^\top\right)\\ &+\trace\left(\Rphi\C\Rpsi^\top\Lambdacols\Rpsi\C^\top\Rphi^\top\right).
    \end{split}
\end{equation}

\subsection{Extensions}
\paragraph{Additional regularization via spectral filtering.}
We propose a stronger explicit regularization by demanding that both $\Mhat$ and $\Mlatent$ be smooth on their respective graphs. Since we do not know the Laplacian of $\Gprime$, we smooth $\Mlatent$ via \textit{spectral filtering}, i.e., through direct manipulation of its spectral representation $\C$. To that end, we pass $\C$ through a bank of pre-chosen spectral filters $\left\{\Frows\right\}_{p\in\mathcal{P}},\left\{\Fcols\right\}_{q\in\mathcal{Q}}$, i.e., diagonal positive semi-definite matrices, and transport the filtered signals to $\Gprod$ according to
\begin{equation}\label{eq:spectralFilter}
    \begin{split}
    &\Mhat_{p,q}=\Phibold\Rphi\Frows \C\Fcols^\top\Rpsi^\top\Psibold^\top,\;\;p\in \mathcal{P},q\in \mathcal{Q}.
    \end{split}
\end{equation}
In particular, we use the following filters,
\begin{equation}
        \Frows = \mathrm{diag}\left(\bb{1}_p\right),\; \Fcols = \mathrm{diag}\left(\bb{1}_q\right),\;
\end{equation}
where $\bb{1}_k=\left[\begin{array}{ccccccc}1&\ldots&1&0&\ldots&0 \end{array}\right]^\top$ denotes a vector with $k$ ones followed by zeros.
For these manipulations to take effect, we replace $\Ez$ in \eqref{eq:DMFOpt_spect} with the following loss function,
\begin{align}\label{eq:fullzoomoutLoss}
\begin{split}
        \Ezoomout\left(\Rphi,\Mlatent,\Rpsi\right) & = \sum_{\substack{p\in \mathcal{P}\\q \in\mathcal{Q}}}\|\left(\Mhat_{p,q}-\M\right)\odot \mask\|_F^2.
\end{split}
\end{align}
Despite the fact that we used separable filters in \eqref{eq:spectralFilter}, these filters are coupled through the loss \eqref{eq:fullzoomoutLoss}. This results in an overall inseparable spectral filter that still retains a DMF structure, since \eqref{eq:spectralFilter} is a $5$-layer DMF with two fixed layers. While \Cref{thm:sing_vals_evolve} does not cover the case of a multi-layer DMF where only a subset of the layers are trainable, our empirical evaluations suggest that the implicit rank regularization is still in place. This additional regularization allows us to get decent reconstruction errors even when the number of measurements is extremely small, as we show in \Cref{subsec:CSA}.
\paragraph{Regularization of the individual layers.}
Another extension we explore is imposing further regularization on the individual layers. For example, one could ask $\Lrows^\everymodeprime$ and $\A^\top\Lrows\A$ to be jointly diagonalized by $\Phiprime$. Using \eqref{eq:Zspect}-\eqref{eq:Aspect} we get,
\begin{equation}
\Lambdarows^\everymodeprime=\Phiprime^\top\A^\top\Lrows\A\Phiprime=\Rphi
^\top\Phibold^\top\Lrows\Phibold\Rphi=\Rphi^\top\Lambdarows\Rphi.
\end{equation}
Thus, we can approximately enforce this constraint with the following penalty term,
\begin{equation}\label{eq:diag_rows}
    \Ediagrows\left(\Rphi\right) = \|\mathrm{off}\left(\Rphi^\top\Lambdarows\Rphi\right)\|_F^2,
\end{equation}
where $\mathrm{off}(\cdot)$ denotes the off-diagonal elements. A similar treatment to the columns graph gives,
\begin{equation}\label{eq:diag_cols}
    \Ediagcols\left(\Rpsi\right) = \|\mathrm{off}\left(\Rpsi^\top\Lambdacols\Rpsi\right)\|_F^2.
\end{equation}
While these penalty terms are not a function of the product matrix, their inclusion did not harm the implicit regularization.

\section{Experimental study on synthetic data}\label{sec:exp_study}

\begin{figure*}[!t]
\centering
\vfill
\subfigure{\includegraphics[width=0.31\textwidth]{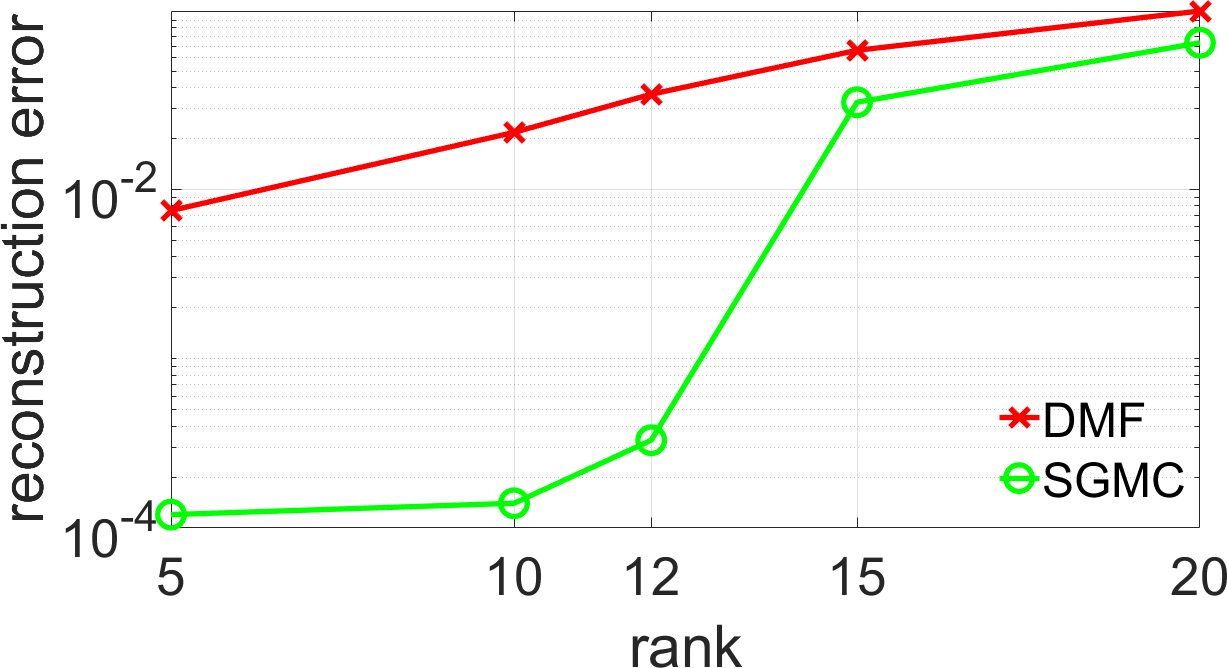}}
\subfigure{\includegraphics[width=0.31\textwidth]{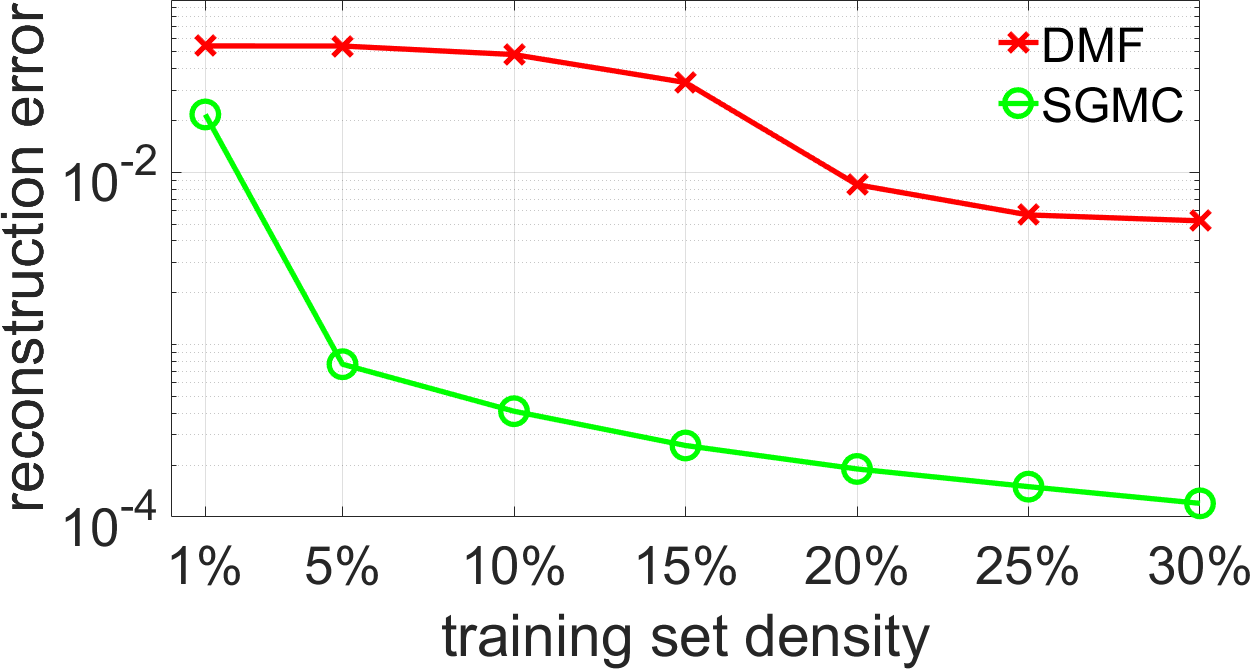}}
\subfigure{\includegraphics[width=0.31\textwidth]{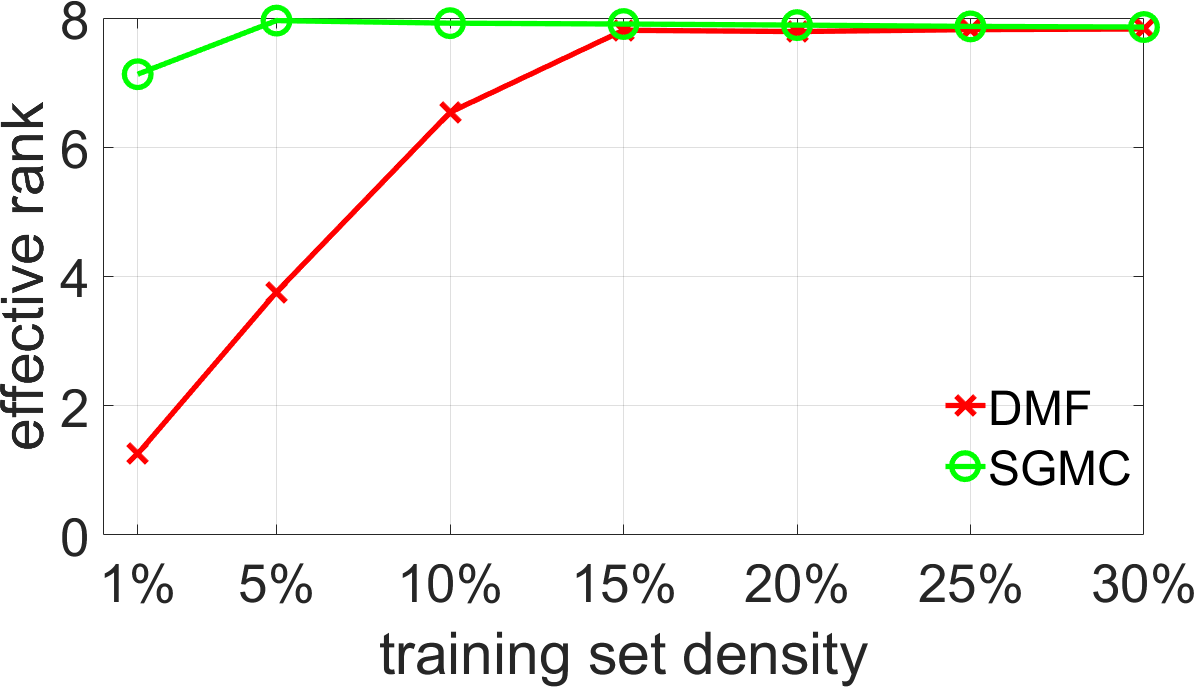}}
\\
\caption{\small In these experiments we generated band-limited (low-rank and smooth) matrices using the synthetic Netflix graphs (see \Cref{fig:synth_netflix}) to test the dependence of SGMC and DMF on the rank of the underlying matrix and on the number of training samples. \textbf{Left}: reconstruction error (on the test set) vs. the rank of the ground-truth matrix. As the rank increases, the reconstruction error increases, but it increases slower for SGMC than for DMF. For the training set we used $15\%$ of the points chosen at random (same training set for all experiments). $\mu$ was set to $0.001$. \textbf{Middle}: reconstruction error (on the test set) vs. density of the sampling set in $\%$ of the number of matrix elements, for a random rank $10$ matrix of size $150 \times 200$. As we increase the number of samples, the gap between DMF and SGMC reduces. Still, even when using $30\%$ of the samples, SGMC performs better for the same number of iterations. For all the experiments we set $\mu=0.01$, $lr=0.001$, $\texttt{maxiter=} 3\times 10^6$. \textbf{Right}: effective rank \citep{roy2007effective} vs. training set density, for a random rank $10$ matrix. Even for extremely data-poor regimes, SGMC was able to recover the effective rank of the ground-truth matrix, whereas is underestimating it.
}
\label{fig:rank_and_sampling}
\end{figure*}

The goal of this section is to compare between our approach and vanilla DMF on a simple example of a community structured graph. We exhaustively compare between the following distinct methods:
\begin{itemize}
    \item \textbf{Deep matrix factorization (DMF):} 
    \begin{equation}\label{eq:dmf_obj}
    \min_{\Rphi,\C,\Rpsi} \left\|\left(\Rphi\C\Rpsi^\top- \M\right) \odot \mask\right\|_F^2,
    \end{equation}
    \item \textbf{Spectral geometric matrix completion (SGMC):} The proposed approach defined by the optimization problem \eqref{eq:DMFOpt_spect}.
    \item \textbf{Functional Maps (FM, SGMC1):} This method is like SGMC with a single layer, i.e., we optimize only for $\C$, while $\Rphi$ and $\Rpsi$ are set to identity.
\end{itemize}
Since SGMC uses additional information, it is expected to perform better than DMF. However, proper utilization of the graph information is not trivial (as is evident by \Cref{table:results}), and we conduct this set of controlled experiments to attest for it.

We use the graphs taken from the synthetic Netflix dataset. Synthetic Netflix is a small synthetic dataset constructed by \cite{kalofolias2014matrix} and \cite{monti2017geometric}, in which the user and item graphs have strong communities structure. See \Cref{fig:synth_netflix} in \Cref{sec:appendixA} for a visualization of the user/item graphs. It is useful in conducting controlled experiments to understand the behavior of geometry-exploiting algorithms. In all our tests we use a randomly generated band-limited matrix on the product graph $\Gcols\squarespace\Grows$. For the complete details please refer to the captions of the relevant figures. 

\paragraph{Performance evaluation.}
To evaluate the performance of the algorithms in this section, we report the \textit{root mean squared error},
\begin{equation}\label{eq:RMSE}
    \mathrm{RMSE}(\Mhat,\mask) = \sqrt{\frac{\left\|\left(\Mhat-\M\right)\odot\mask\right\|_F^2}{\sum_{i,j}\mask_{i,j}}}
\end{equation}
computed on the complement of the training set. Here $\bb{X}$ is the recovered matrix and $\mask$ is the binary mask representing the support of the set on which the RMSE is computed.\\

We explore the following aspects:

\begin{figure*}
    \centering
    \includegraphics[width=0.32\textwidth]{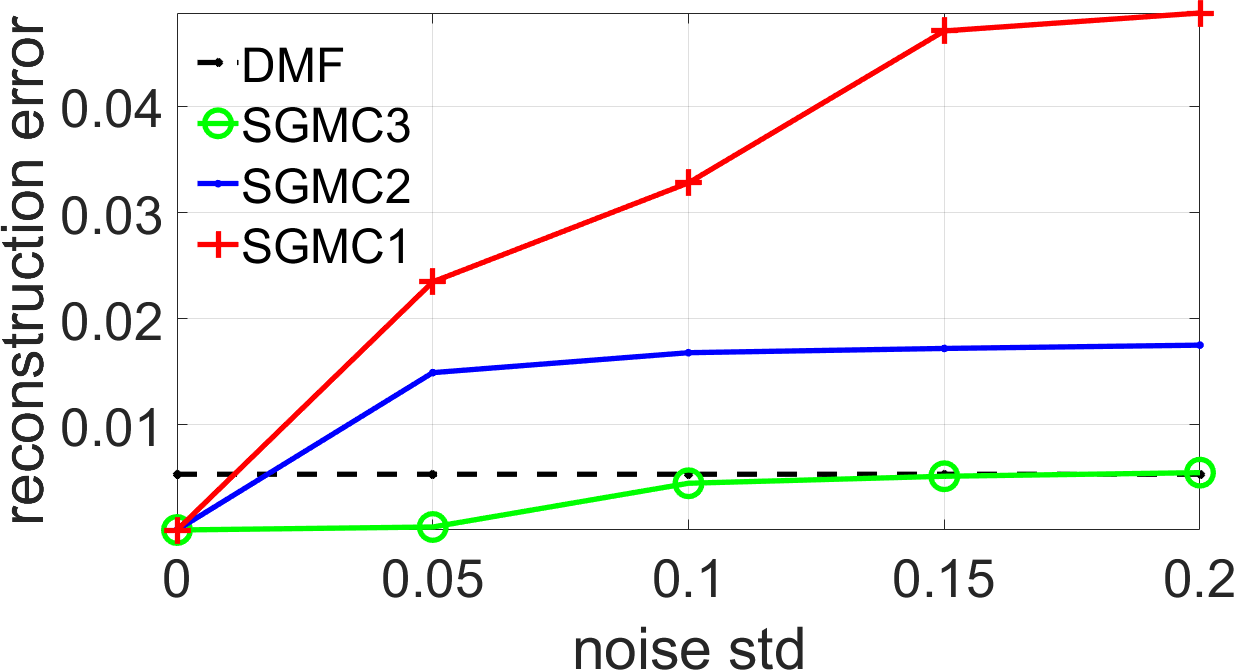}
    \includegraphics[width=0.32\textwidth]{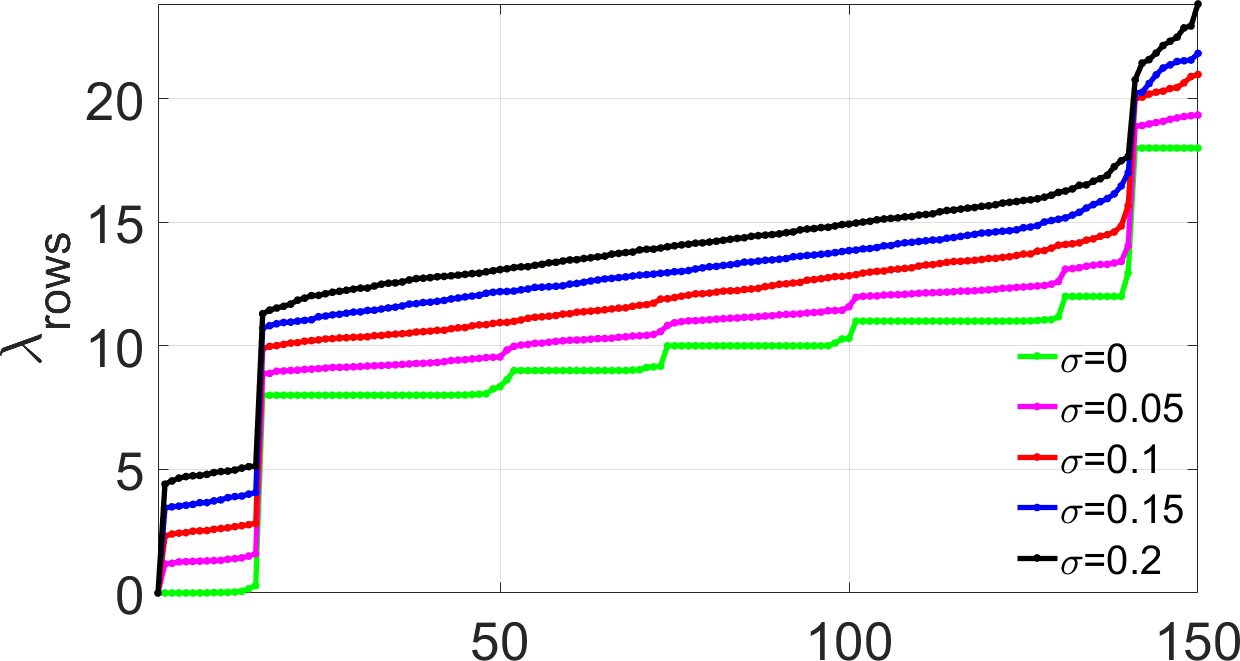}
    \includegraphics[width=0.32\textwidth]{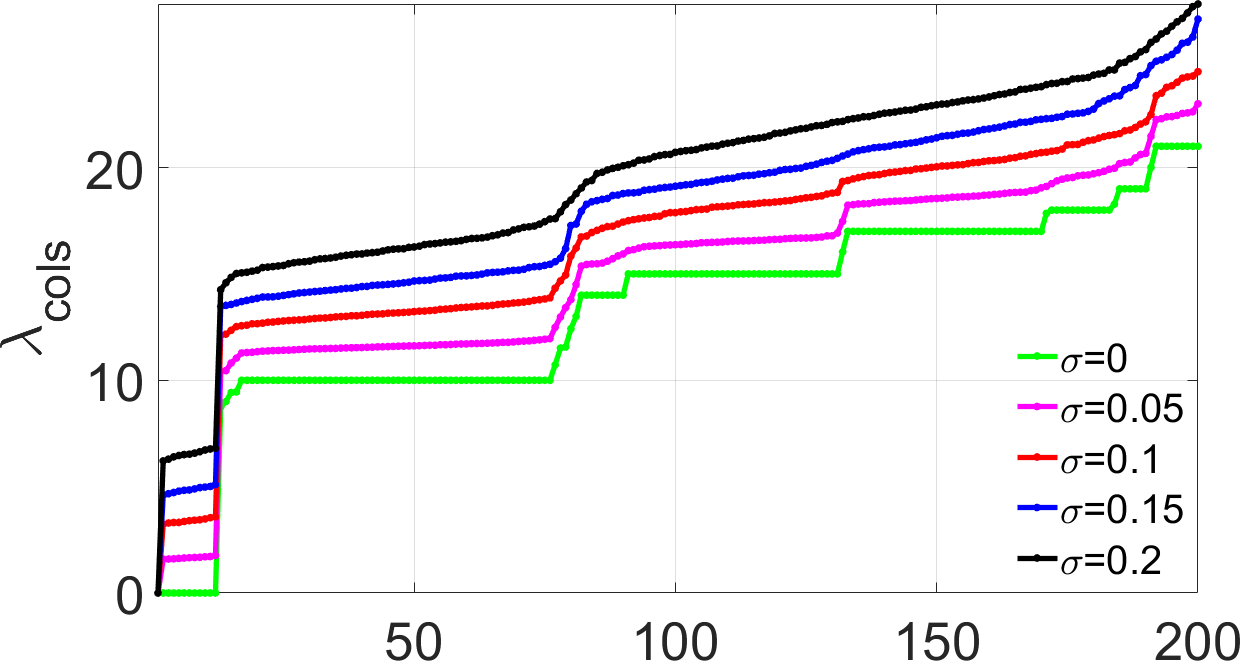}
    \caption{\small In this experiment, we study the robustness of SGMC in the presence of noisy graphs. We perturbed the edges of the graphs by adding random Gaussian noise with zero mean and tunable standard deviation to the adjacency matrix. We discarded the edges that became negative as a result of the noise, and symmetrized the adjacency matrix. SGMC1/SGMC2/SGMC3 stand for SGMC with 1 layer (training only $\C$), 2 layers (training $\C, \Rphi$) and 3 layers ($\C, \Rphi, \Rpsi$). \textbf{Left}: With clean graphs all SGMC methods perform well. As the noise increases, the regularization induced by the depth kicks in and there is a clear advantage for SGMC3. For large noise, SGMC3 and DMF achieve practically the same performance. \textbf{Middle \& Right}: eigenvalues of $\Lrows,\Lcols$ for different noise levels. Even for moderately large amounts of noise, the structure of the lower part of the spectrum is preserved, and the effect on the low-frequency (smooth) signal remains small.}
    \label{fig:noisy_graphs}
\end{figure*}

\begin{figure*}[t]
   \centering
\begin{tabular}{c c c c}
\addtolength{\tabcolsep}{-20pt} 
& \large $1\%$  & \large $5\%$ & \large $10\%$   \\
\rotatebox[origin=c]{90}{\large \;\;\;\;\;\;\;\;\;\;\;\;\;\;\;\;\;\;\;\;\;\;\;\;\;\;\;\;\;\;\;\;  DMF} & {
\includegraphics[width=0.26\textwidth]{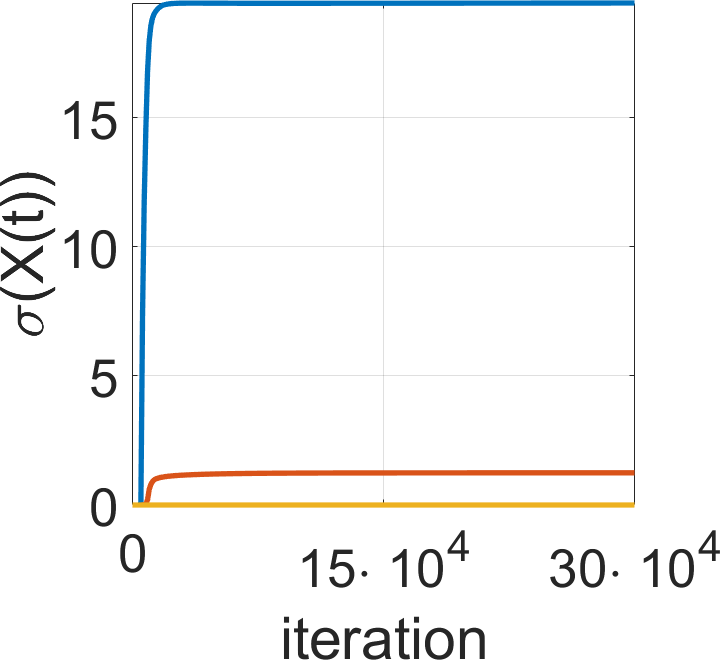}} &
{\includegraphics[width=0.26\textwidth]{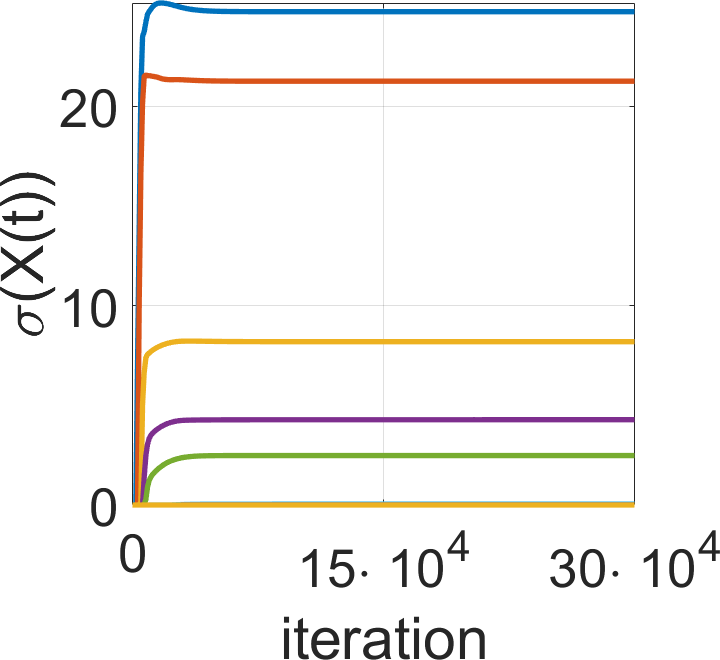}} &
{\includegraphics[width=0.26\textwidth]{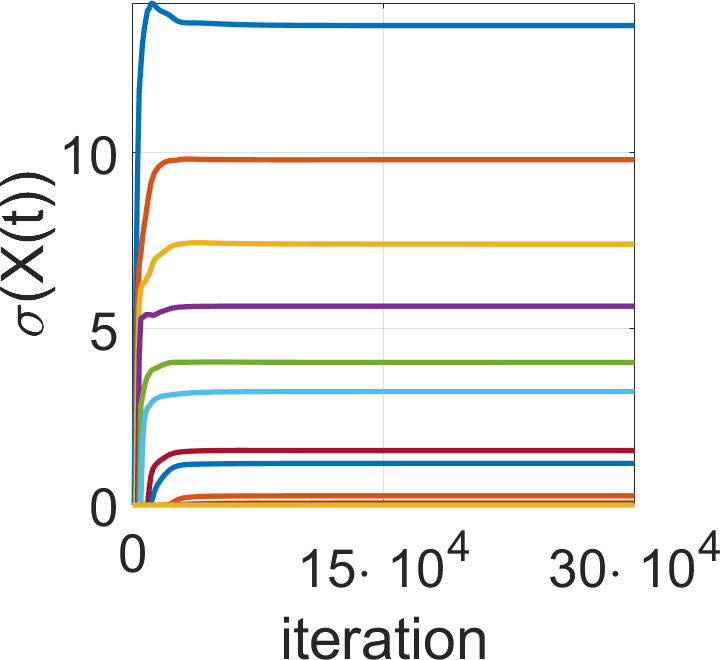}} 
 \vspace{-2.1cm}
 \\
\rotatebox[origin=c]{90}{\large\;\;\;\;\;\;\;\;\;\;\;\;\;\;\;\;\;\;\;\;\;\;\;\;\;\;\;\;\;\;\;\;  SGMC} & {
\includegraphics[width=0.26\textwidth]{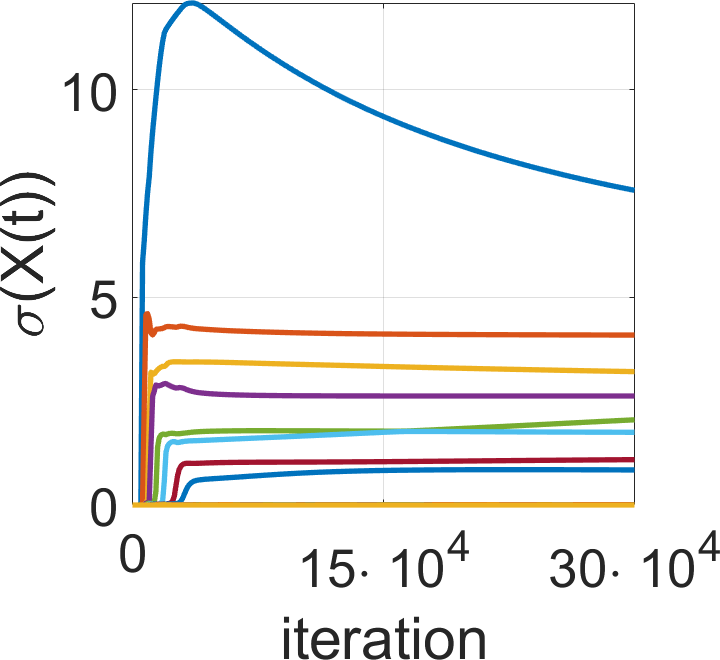}} &
{\includegraphics[width=0.26\textwidth]{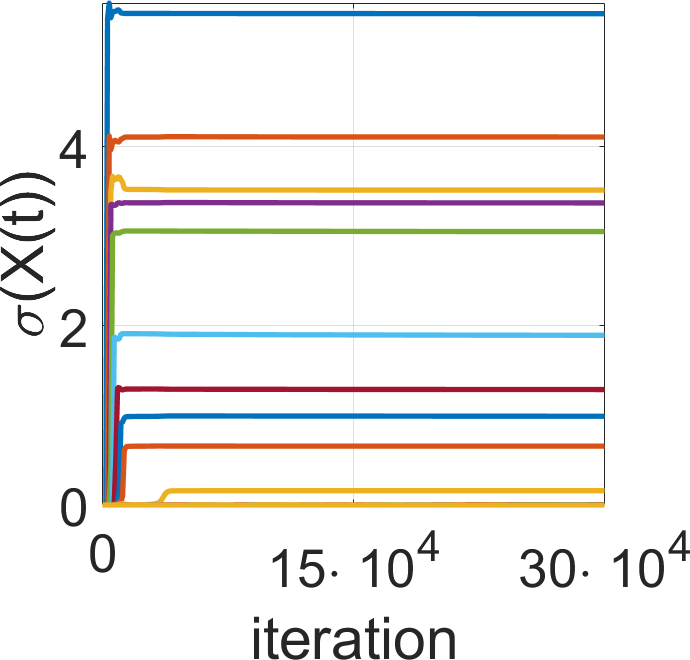}} &
{\includegraphics[width=0.26\textwidth]{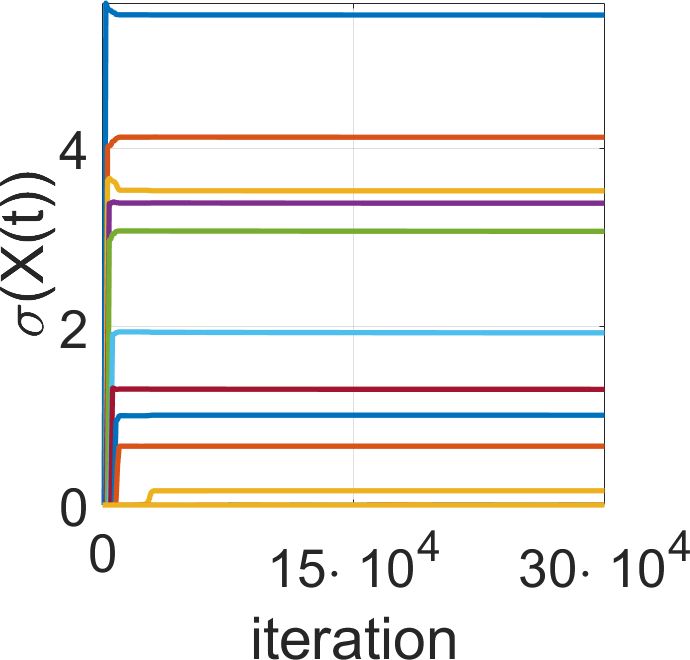}} 
\end{tabular}  
 \vspace{-2.2cm}
\caption{\small In these experiments, we plot the dynamics of the singular values of the product matrix $\Mhat(t)$ during the gradient descent iterations. 
We show singular value convergence at different sampling densities (left-to-right: $1\%$, $5\%$, and $10\%$) for SGMC and DMF. We use the synthetic Netflix graphs on which we generate a random rank $10$ matrix of size $150 \times 200$.
In accordance with \Cref{fig:rank_and_sampling}, we see that SGMC is able to recover the rank even for a very data-poor regime, whereas DMF demands significantly higher sample complexity.
\label{fig:sing_val_cvg}}
\end{figure*}
\paragraph{Sampling density.}
We investigate the effect of the number of samples on the reconstruction error and the \textit{effective rank} of the recovered matrix \citep{roy2007effective}.
We demonstrate that in the data-poor regime, the implicit regularization of DMF is too strong resulting in poor recovery, compared to a superior performance achieved by incorporating geometric regularization through SGMC. These experiments are summarized in \Cref{fig:rank_and_sampling}.
\paragraph{Initialization.}
In all of our experiments we initialize with balanced initialization \eqref{eq:balanced}, with scaled identity matrices $10^{-\alpha} \bb{I}$. We explore the effect of initialization in \Cref{fig:init} (in \Cref{sec:appendixA}).
\paragraph{Rank of the underlying matrix.}
We explore the effect of the rank of the underlying matrix, showing that as the rank increases it becomes harder for both SGMC and DMF to recover the matrix. A remarkable property of SGMC is that it is able to get a decent approximation of the effective rank of the matrix even with extremely low number of samples. These  experiments are summarized in \Cref{fig:rank_and_sampling}.
\paragraph{Noisy graphs.}
We study the effect of noisy graphs on the preformance of SGMC. \Cref{fig:noisy_graphs} demonstrates that SGMC is able to utilize graphs with substantial amounts of noise before its performance drops to the level of vanilla DMF (which does not rely on any knowledge of the row/column  graphs).
\paragraph{Dynamics.}
\Cref{fig:sing_val_cvg} shows the dynamics of the singular values of $\Mhat$ during the optimization. We visually verify that they behave according to \eqref{eq:S_evolve} and that the convergence rate of the relevant singular values is much faster in SGMC than in DMF.
\paragraph{Code.}
An interactive jupyter notebook is available \href{https://colab.research.google.com/drive/1OkNEiTHok14gcVf3NxFIbAFutDN6-Tx6?usp=sharing}{here}.
\section{Results on recommender systems datasets}
We demonstrate the effectiveness of our approach on the following datasets: Synthetic Netflix, Flixster, Douban, Movielens (ML-100K) and Movielens-1M (ML-1M) as referenced in \Cref{table:results}. The datasets include user ratings for items (such as movies) and additional features. For all the datasets we use the users and items graphs taken from \href{https://github.com/fmonti/mgcnn}{\citet{monti2017geometric}}. The ML-1M dataset was taken from \href{https://github.com/riannevdberg/gc-mc }{\citet{berg2017graph}}, for which we constructed 10 nearest neighbor graphs for users/items from the features, and used a Gaussian kernel with $\sigma=1$ for edge weights. See \Cref{table:datasets} in \Cref{sec:appendixA} for a summary of the dataset statistics. For all the datasets, we report the results for the same test splits as that of \citet{monti2017geometric} and \citet{berg2017graph}.
The compared methods are referenced in \Cref{table:results}.

\paragraph{Proposed baselines.} We report the results obtained using the methods discussed above, with the addition of the following method:
\begin{itemize}
    \item \textbf{SGMC-Z}: a variant of SGMC that uses \eqref{eq:fullzoomoutLoss} as a data term.
    For this method we chose a maximal value of $\pmax,\qmax$ (which can be larger than $m,n$) and a skip determining the spectral resolution, denoted by $\pskip,\qskip$. We use $p=1+k\pskip,q=1+k\qskip,\;k\in\mathbb{N}$.
\end{itemize}

In addition, we add the diagonalization terms \eqref{eq:diag_rows}, \eqref{eq:diag_cols} weighted by $\diagweight_r,\diagweight_c$, respectively, to the SGMC/SGMC-Z methods. The optimization is carried out using gradient descent with fixed step size (i.e., fixed learning rate), which is provided for each experiment alongside all the other hyper-parameters in \Cref{table:params}.

\paragraph{Initialization.} 
All our methods are deterministic and did not require multiple runs to account for initialization. 
We always initialize the matrices $\Rphi,\C, \Rpsi$ with $10^{-\alpha}\bb{I}$. In \Cref{fig:init} we reported results on synthetic Netflix and ML-100K datasets for different values of $\alpha$. We noticed that for SGMC and SGMC-Z it is best to use $10^{-\alpha}=1$. According to \citep{gunasekar2017implicit,li2017algorithmic}, DMF requires a large $\alpha$ to decrease the generalization error. 
We used DMF with $10^{-\alpha}=0.01$ for Synthetic Netflix and $10^{-\alpha}=1$ for the real world datasets, in accordance with \Cref{fig:init} and our experimentation.
In the cases where only one of the bases was available, such as in Douban and Flixster-user only benchmarks, we set the basis corresponding to the absent graph to identity. 

\paragraph{Stopping condition.}
Our stopping condition for the gradient descent iterations is based on a validation set. We use $95\%$ of the available entries for training (i.e., to construct the mask $\mask$) and the rest $5\%$ for validation. The $95/5$ split was chosen at random. We stop the iterations when the RMSE \eqref{eq:RMSE}, evaluated on the validation set, does not change by more than $\texttt{tol}=0.000001$ between two consecutive iterations, $|\mathrm{RMSE}_{k}-\mathrm{RMSE}_{k-1}|<\texttt{tol}$. 
Since we did not apply any optimization into the choice of the validation set, we also report the best RMSE achieved on the test set via early stopping. In this regard, the number of iterations is yet another hyper parameter that has to be tuned for best performance.

\subsection{Cold start analysis}
\label{subsec:CSA}
\begin{wrapfigure}{r}{0.6\textwidth}
\vspace{-1.0cm}
\includegraphics[width=0.6\textwidth]{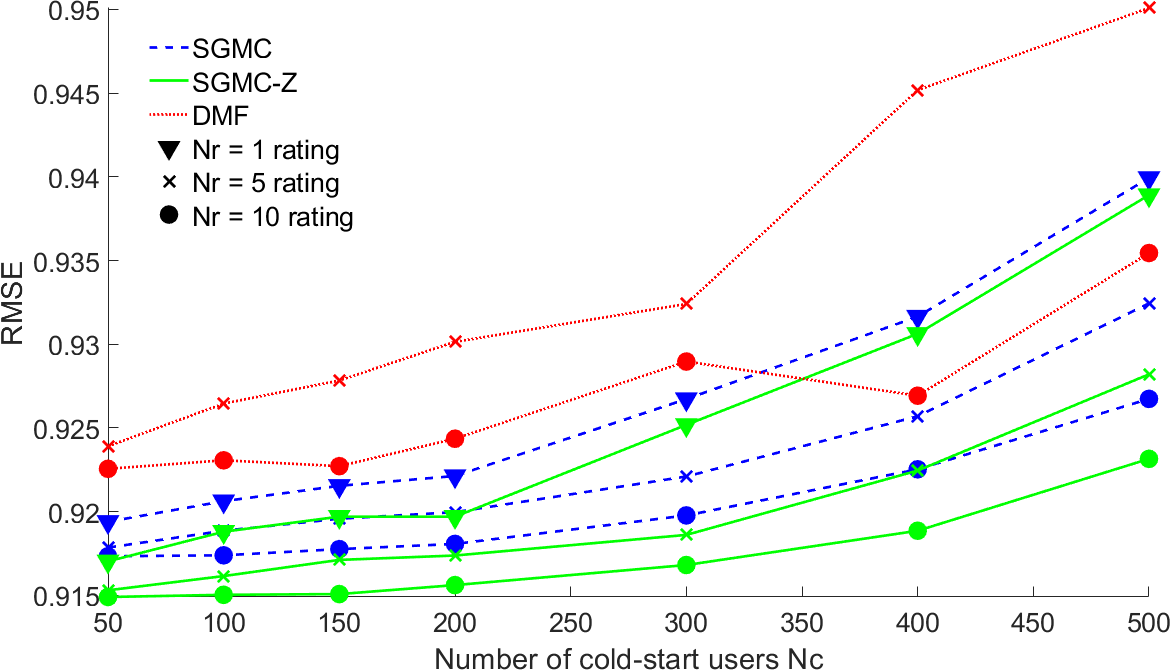}
\captionsetup{labelformat=empty}
\caption{\small Comparison of test RMSE in the presence of cold start users on the ML-100K dataset. The x-axis corresponds to the number of the cold start users $N_c = 50, 100, \ldots 500$. Red, blue and green correspond to DMF, SGMC and SGMC-Z methods respectively as also shown in the legend. Different shapes of the markers indicate different number of maximum ratings ($N_r = \{1, 5, 10\}$) available per cold-start user.}
\vspace{-0.8cm}
\end{wrapfigure}
A particularly interesting scenario in the context of recommender systems is the presence of \textit{cold-start} users, referring to the users who have not rated enough movies yet. We perform an analysis of the performance of our method in the presence of such cold start users on the ML-100K dataset. In order to generate a dataset consisting of $N_c$ cold start users, we sort the users according to the number of ratings provided by each user, and retain at most $N_r$ ratings (chosen randomly) of the bottom $N_c$ users (i.e., the users who provided the least ratings). We choose the values $N_c = \{ 50, 100, 150, 200, 300, 400, 500 \}$ and $N_r = \{1, 5, 10\}$, and run our algorithms: DMF, SGMC and SGMC-Z, with the same hyperparameter settings used for obtaining \Cref{table:results}. We use the official ML-100K test set for evaluation. Similar to before, we use $5\%$ of the training samples as a validation set used for determining the stopping condition.  

The results presented in the inline figure suggest that the SGMC and SGMC-Z outperform DMF significantly, indicating the importance of the geometry as data becomes scarcer. As expected, we can see that the performance drops as the number of ratings per user decreases. Furthermore, we can observe that SGMC-Z consistently outperforms SGMC by a small margin. We note that SGMC-Z, even in the presence of $N_c = 500$ cold start users with $N_r = 5$ ratings, is still able to outperform the full data performance of \citet{monti2017geometric}, demonstrating the strength of geometry and implicit low-rank induced by SGMC-Z. 


\begin{table*}
\centering
\begin{threeparttable}[!t]
\begin{tabular}{l r r r r r r r}
\textbf{Model} & $\substack{ \textbf{Synthetic}\\\textbf{Neflix}}$& \textbf{Flixster} & \textbf{Douban}  & \textbf{ML-100K}&
\\[0.05em] \\[-0.8em]
MC \citep{candes2009exact}& -- & $1.533$ & $0.845$ & $0.973$ \\
GMC \citep{kalofolias2014matrix} & $0.3693$ & -- & -- & $0.996$ \\
GRALS \citep{rao2015collaborative} & $0.0114$ & $1.313/1.245$ & $0.833$ & $0.945$  \\
RGCNN \citep{monti2017geometric} & $0.0053$\tnote{a} & $1.179/0.926$ & $0.801$ & $0.929$  \\
GC-MC \citep{berg2017graph} & -- & $\mathbf{0.941}/{0.917}$ & ${0.734}$ & ${0.910}$\tnote{b} \\
FM (ours) & $0.0064$ &  $3.32$ & $3.15$ & $1.10$  \\
DMF \citep{arora2019implicit}, (ours) & $0.0468$\tnote{d} & $1.06$ & $0.732$ & ${0.918}$\tnote{c} / $0.922$  \\
SGMC (ours)& $\mathbf{0.0021}$ & $0.971$ / $0.900$ & $\mathbf{0.731}$ & $0.912$  \\
SGMC-Z (ours) & $0.0036$ & $0.957$ / $\mathbf{0.888}$ & ${0.733}$ & $\mathbf{0.907}$\tnote{c} / ${0.913}$ \\
\end{tabular}
\begin{tablenotes}
    \item[a]{\scriptsize This number corresponds to the inseparable version of MGCNN.
    \item[b]{\scriptsize This number corresponds to GC-MC.}
    \item[c]{\scriptsize Early stopping.}
    \item[d]{\scriptsize Initialization with $0.01\mathbf{I}$.}} 
\end{tablenotes}
\vspace{-0.2cm}
\caption{\small RMSE test set scores for runs on Synthetic Netflix \citep{monti2017geometric}, Flixster \citep{jamali2010matrix}, Douban \citep{ma2011recommender}, and Movielens-100K \citep{harper2016movielens}. For Flixster, we show results for both user/item graphs (right number) and user graph only (left number). Baseline numbers are taken from \citep{monti2017geometric,berg2017graph}.
\label{table:results}} 
\end{threeparttable}
\vspace{-0.6cm}
\end{table*}


\paragraph{Scalability.}
All the experiments presented in the paper were conducted on a machine consisting of 64GB CPU memory, on an NVIDIA GTX 2080Ti GPU. Most of our large-scale experiments take upto 10-30 minutes of time until convergence, therefore, are rather quick. In this work we focused on the conceptual idea of solving matrix completion via the framework of deep matrix factorization by incorporating geometric regularization, paying little attention to the issue of scalability. There are two main computational bottlenecks to our approach: The spatial version \eqref{eq:DMFOpt} requires the computation of the matrix product $\A\Mlatent\B^\top$ in each gradient iteration, and the spectral version requires also the
eigenvalue decomposition of $\Lrows,\Lcols$. 
These limitations apply to other graph neural networks as well \citep{hu2020open}, and we believe that they can be at least partially addressed by ad-hoc solutions.

\subsection{Discussion}

A few remarkable observations can be extracted from \Cref{table:results}:
First, on the Douban and ML-100K dataesets, vanilla DMF shows competitive performance with all the other methods. This suggests that the geometric information is not very useful for these datasets.
Second, the proposed SGMC algorithms outperform the other methods, despite their simple and fully linear architecture.
This suggests that the other geometric methods do not exploit the geometry properly, and this fact is obscured by their cumbersome architecture.
Third, while some of the experiments reported in \Cref{table:results} showed only slight margins in favor of SGMC/SGMC-Z compared to DMF, the results in the Synthetic Netflix column, the ones reported on Synthetic Movielens-100K (\Cref{tab:results_synthetic_movielens} in \Cref{sec:appendixA}) and the ones reported in \Cref{fig:rank_and_sampling}, suggest that when the geometric model is accurate our methods demonstrate superior results. \Cref{tab:ML1M} in \Cref{sec:appendixA} presents the results of Movielens-1M. First, we can deduce that vanilla DMF model is able to  match the performance of complex alternatives. Furthermore, using graphs produces slight improvements over the DMF baseline and overall provides competitive performance compared to heavily engineered methods. On Synthetic Netflix, we notice that by using SGMC, we outperform \citet{monti2017geometric} by a significant margin, reducing the test RMSE by half. Additionally, it can be observed that DMF performs poorly on both synthetic datasets compared to SGMC/SGMC-Z, raising a question as to the quality of the graphs provided with those datasets on which DMF performed comparably. 

A compelling argument for this behaviour is given by \Cref{table:datasets} in \Cref{sec:appendixA}. We can see that in the real datasets we tested on, the number of available samples is way below the density required by DMF to achieve good performance, in accordance with our findings in \Cref{sec:exp_study}. With high quality graphs, we should have expected SGMC to outperform DMF by a large margin.

\section{Results on drug-target interaction}
In this section we demonstrate the effectiveness of our approach on the problem of predicting drug-target interaction (DTI). The task is to find effective interactions between chemical compounds (drugs)
and amino-acid sequences/proteins (targets). This is traditionally done through wet-lab
experiments which are costly and laborious, and lead to high attrition rate. One possible way to improve this procedure is to predict interaction probabilities through a computational model. To that end, DTI can be interpreted as a matrix completion problem where the rows correspond to different drugs and the columns correspond to different targets. Each entry in the matrix corresponds to the probability of interaction between a drug and target. 
We assume that we are given two graphs encoding similarities between drugs and similarities between targets. The similarity between two drugs is measured by the number of shared substructures within their chemical structures. The similarity between targets is given by their genomic sequence similarity. These similarity measures constitute a standard similarity score that is common in the DTI prediction task.
For more information on the problem and the construction of the graphs, we refer to \citet{mongia2020drug} and references therein. 

\paragraph{Datasets.}
We use three benchmark datasets introduced in \citet{yamanishi2008prediction}, having three different classes of proteins:
enzymes (Es), ion channels (ICs), and G protein-coupled receptors (GPCRs). The data was simulated from public databases KEGG BRITE \citep{kanehisa2006genomics}, BRENDA
\citep{schomburg2004brenda} SuperTarget \citep{gunther2007supertarget} and DrugBank \citep{wishart2008drugbank}, and is publicly available\footnote{\href{http://web.kuicr.kyoto-u.ac.jp/supp/yoshi/drugtarget/}{http://web.kuicr.kyoto-u.ac.jp/supp/yoshi/drugtarget/}}.
The data from each of these databases is formatted as an adjacency matrix between drugs and targets encoding the interaction as $1$ if drug-target pair are known to interact and $0$ otherwise.
\paragraph{Baselines.}
We validated our proposed method by comparing it with three recent methods proposed in the literature: MGRNNM \citep{mongia2020drug}, GRMF \citep{ezzat2016drug}, CMF \citep{zheng2013collaborative}. For all the baselines we ran the publicly available code\footnote{\href{https://github.com/aanchalMongia/MGRNNMforDTI}{https://github.com/aanchalMongia/MGRNNMforDTI}} on the aforementioned datasets using the same graphs and same train-test splits.
\paragraph{Evaluation protocol.}
Similarly to \citep{mongia2020drug}, we have performed $5$ runs (with different random seeds) of $10$-fold cross-validation for each of the algorithms under three cross-validation settings (CVS):
\begin{itemize}
    \item CVS1/Pair prediction: random drug–target pairs are chosen randomly for the test set. It is the conventional setting for validation and evaluation.
    
    \item CVS2/Drug prediction: complete drug profiles are left out of the training set, i.e., some rows are absent. This tests the algorithm’s ability to predict interactions for novel drugs for which no interaction information is available.

\item CVS3/Target prediction: complete target profiles are left out of the training set. i.e., some columns are absent. It tests the algorithm’s ability to predict interactions for novel targets.
\end{itemize}

Out of the 10 folds one was left out for testing whereas the remaining 9 folds were used as the training set.
To evaluate performence we measure area under ROC curve (AUC), area under the precision-recall curve (AUPR), and RMSE. In biological drug discovery, AUPR is of more significance since it penalizes high ranked false positive interactions much more than AUC. Those pairs would be biologically validated later in the drug discovery process. The results are summarised in \Cref{table:DTI} in \Cref{sec:appendixB}.

\paragraph{Discussion.}
\Cref{table:DTI} clearly shows that SGMC mostly outperforms the other methods in all metrics. This is without applying any particular task specific optimization of the loss function and other hyper-parameters.  In particular, the RMSE criterion, which is the one optimized by all the methods, is significantly lower for SGMC compared to other matrix factorization algorithms. This serves as a further reinforcement of the strength of the implicit regularization in SGMC compared to the nuclear norm \citep{mongia2020drug} and explicit low rank matrix factorization methods \citep{ezzat2016drug}.


\section{Related work}
\paragraph{Geometric matrix completion.}
There is a vast literature on classical approaches for matrix completion, and covering it is beyond the scope of this paper. In recent years,
the advent of deep learning platforms equipped with
efficient automatic differentiation tools allows the exploration of sophisticated models that incorporate intricate regularizations.
Some of these contemporary approaches to matrix completion fall under the umbrella term of \textit{geometric deep learning}, which generalizes standard (Euclidean) deep learning to domains such as graphs and manifolds. For example, \textit{graph convolutional neural networks} (GCNNs) follow the architecture of standard CNNs, but replace the Euclidean convolution operator with linear filters constructed using the graph Laplacian. We distinguish between \textit{inductive} approaches to matrix completion, which work directly on the users and items features to predict the rating matrix (e.g., \citet{berg2017graph}), and \textit{transductive} approaches, which make use of side information to construct graphs encoding relations between rows/columns \citep{kovnatsky2014functional,kalofolias2014matrix,monti2017geometric}.

More recently, it has been demonstrated that some graph CNN architectures can be greatly simplified, and still perform competitively on several graph analysis tasks \citep{wu2019simplifying}. Such simple techniques have the advantage of being easier to analyze and reproduce. One of the simplest notable approaches is \textit{deep linear networks}, networks comprising of only linear layers. While these network are still mostly used for theoretical investigations, we note the recent results of \citet{kliger2019blind} who successfully employed such a network for the tasks of blind image deblurring, and \citep{richardson2020surprising} who used it for image-to-image translation. \citet{jing2020implicit} showed that overparametrized linear layers can be used for implicit rank minimization within a generative model.

A closely related field dealing with reconstruction of signals defined on graphs is graph signal processing. In this field the problem is attacked by extending results from harmonic analysis to problems defined on graphs. For example, \citet{puy2018structured, puy2018random} have developed a random sampling strategy that provides reconstruction guarantees for bandlimited signals on graphs. The reconstruction is performed via minimizing an $l_2$ data term with Dirichlet regularization \eqref{eq:quadform}. Random sampling schemes and reconstruction guarantees for bandlimited signals on product graphs were developed in \cite{ortiz2018sampling, varma2018sampling}. While these results are extremely useful in designing sampling strategies for bandlimited signals on graphs, they are of less use when we are given the samples upfront and have no ability to control the sampling process. Nevertheless, their analysis sheds light on the success of spectral regularization in reconstruction problems on graphs and we intend to integrate these ideas with our approach in the future.

\paragraph{Product manifold filter \& Zoomout.}
The inspiration for our paper stems from techniques for finding shape correspondence. In particular, the functional maps framework and its variants \citep{ovsjanikov2012functional, ovsjanikov2016computing}.  Most notably the work of \citet{litany2017fully} who combined functional maps with joint diagonalization to solve partial shape matching problems, and the \textit{product manifold filter} (PMF) \citep{vestner2017efficient, vestner2017product} and \textit{zoomout} \citep{melzi2019zoomout} -- two greedy algorithms for correspondence refinement by gradual introduction of high frequencies.


\section{Conclusion}
\label{sec:conclusion}
In this work we have proposed a simple spectral technique for matrix completion, building upon recent practical and theoretical results in geometry processing and deep linear networks. We have shown, through extensive experimentation on real and synthetic datasets across domains, that combining the implicit regularization of DMF with explicit, and possibly noisy, geometric priors can be extremely useful in data-poor regimes. Our work is a step towards building interpretable models that are grounded in theory, and proves that such simple models need not only be considered for theoretical study. Through a proper lens, they can be made useful.

\acks{We thank Angshul Majumdar for useful discussions, and for providing datasets and code for the matrix completion methods used in drug-target interaction prediction. This research was supported by ERC StG RAPID and ERC CoG EARS.}

\bibliography{refs}

\appendix

\section{Recommendation systems}
\label{sec:appendixA}

\begin{figure*}[!h]
\centering
\vfill
\subfigure{   \includegraphics[width=0.48\textwidth]{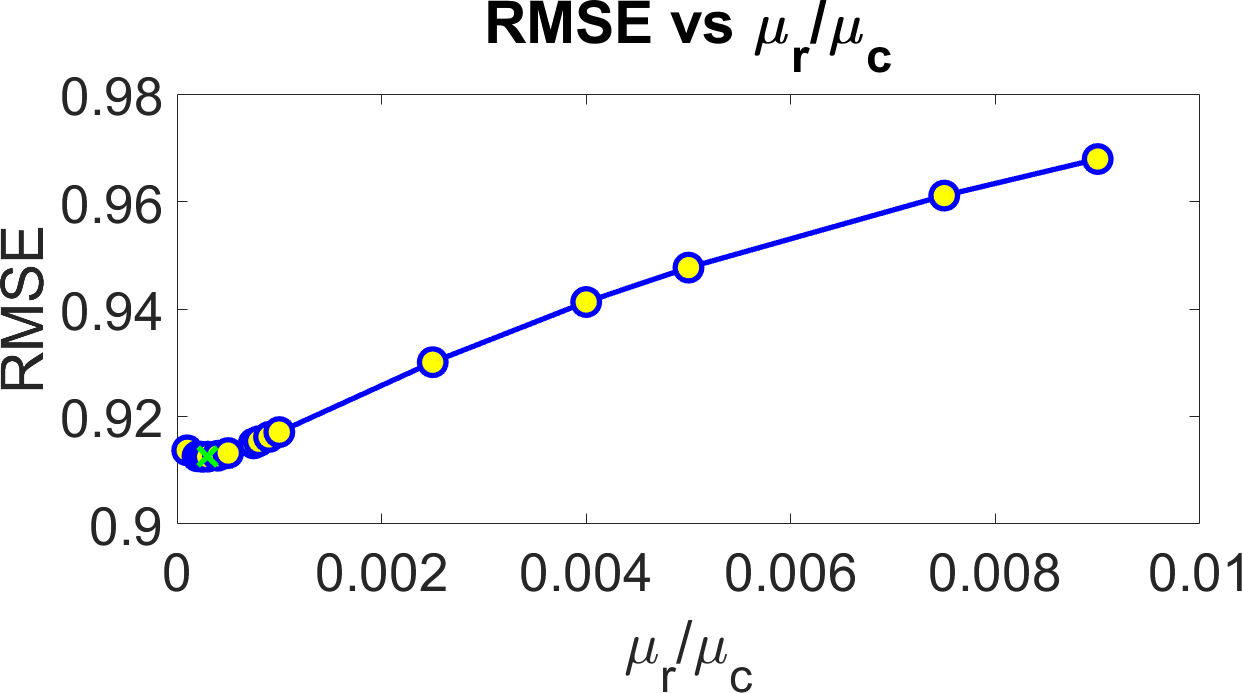}}
 \subfigure{  \includegraphics[width=0.48\textwidth]{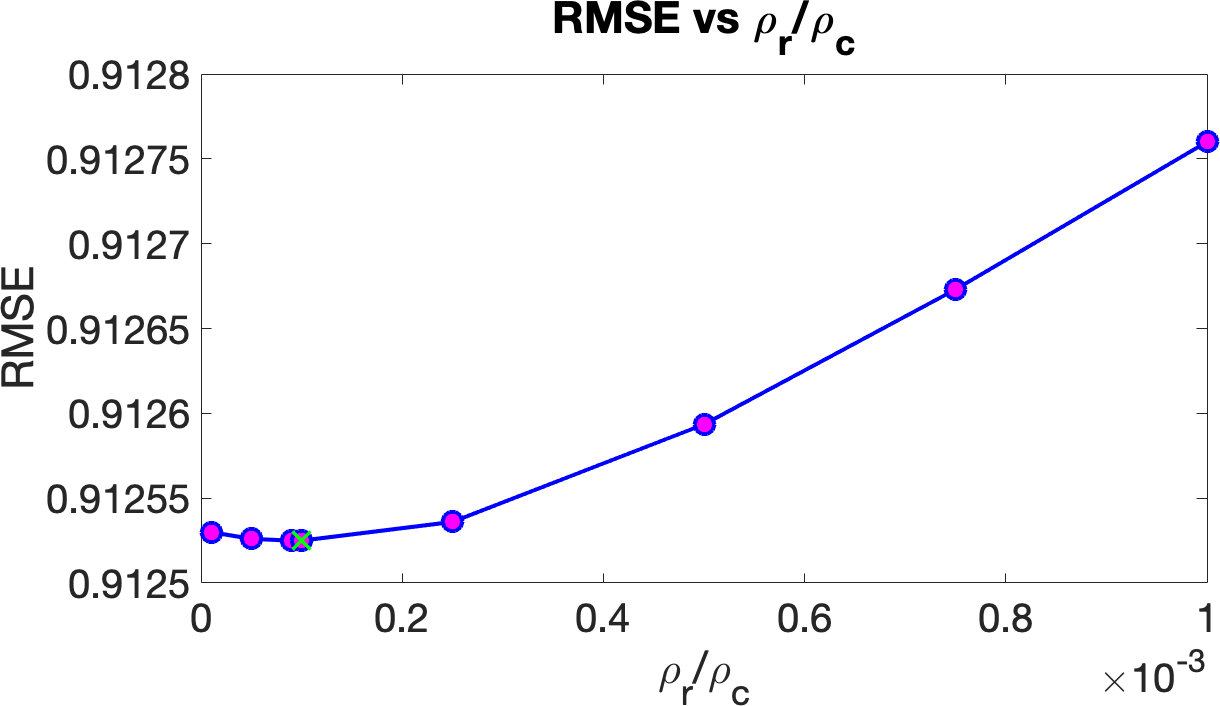}}
\caption{Ablating $\diagweight_r=\diagweight_c$ and $\dirweight_r=\dirweight_c$ of SGMC on the ML-100K dataset. The rest of the parameters were set to the ones reported in \Cref{table:params}. Green X denotes the baseline from \Cref{table:results}.}
\label{fig:ablation_SGMC}
\end{figure*}

\begin{figure*}[!h]
\centering
\vfill
 \subfigure{\includegraphics[width=0.48\textwidth]{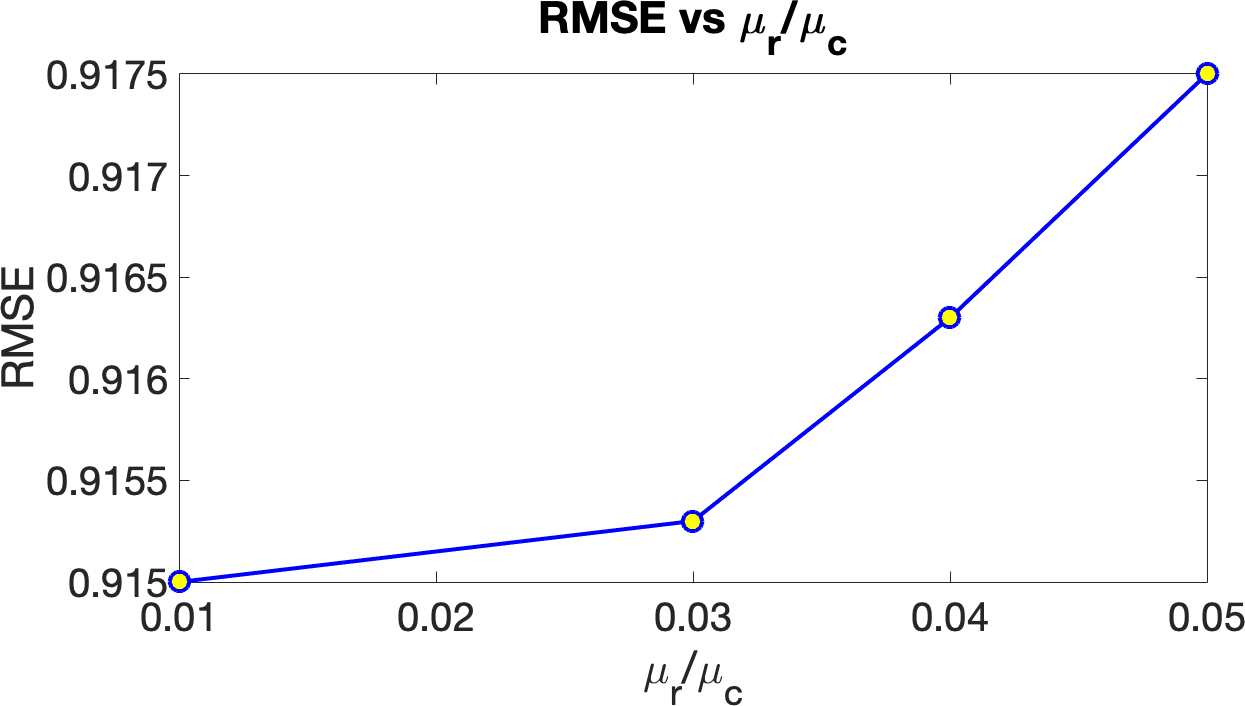}}
 \subfigure{\includegraphics[width=0.48\textwidth]{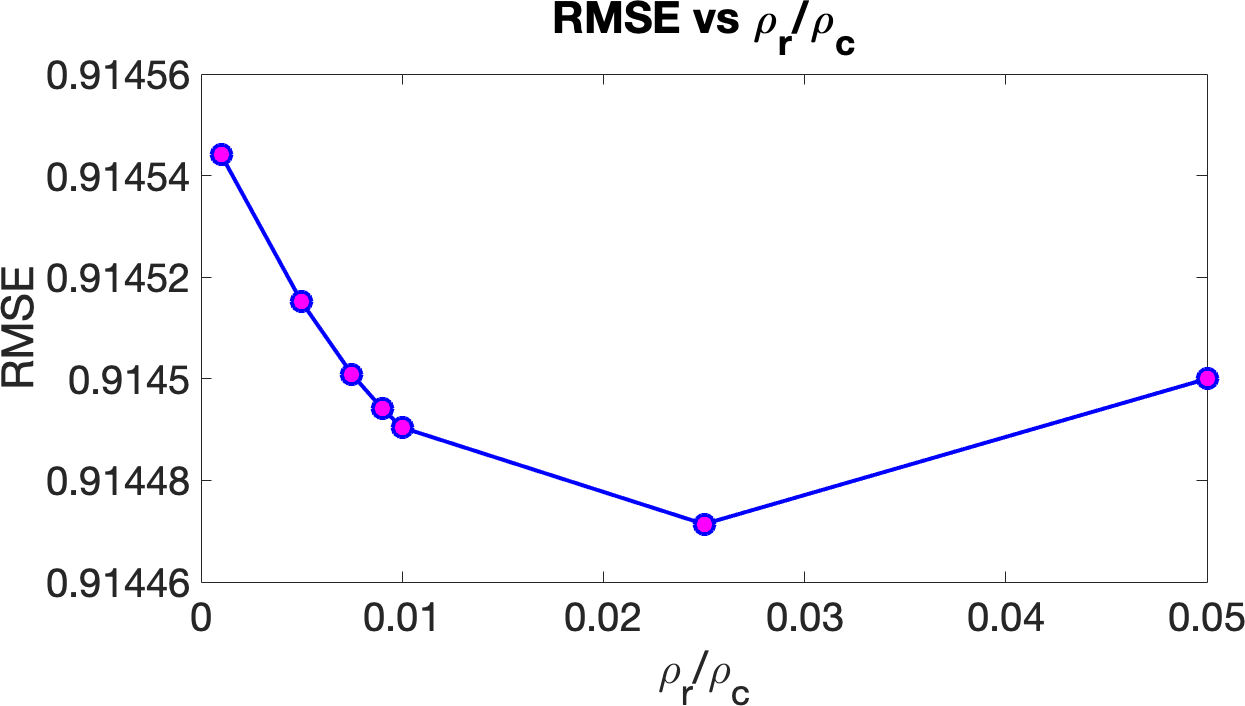}}
\caption{Ablating $\diagweight_r, \diagweight_c$ and $\dirweight_r, \dirweight_c$ of SGMC-Z on the ML-100K dataset. The rest of the parameters were set to the ones reported in \Cref{table:params}.}
\label{fig:ablation_SGMCZ}
\end{figure*}

\begin{figure*}[!h]
\centering
\vfill
\subfigure{\includegraphics[width=0.48\textwidth]{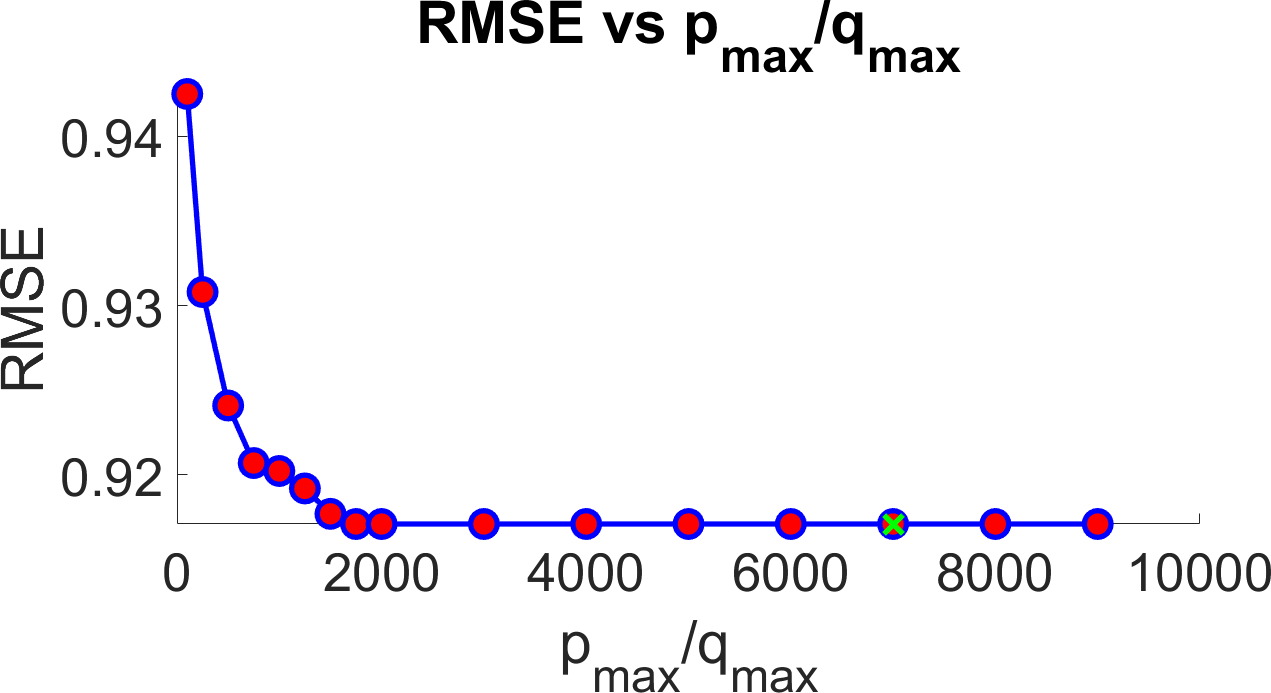}}
\subfigure{\includegraphics[width=0.48\textwidth]{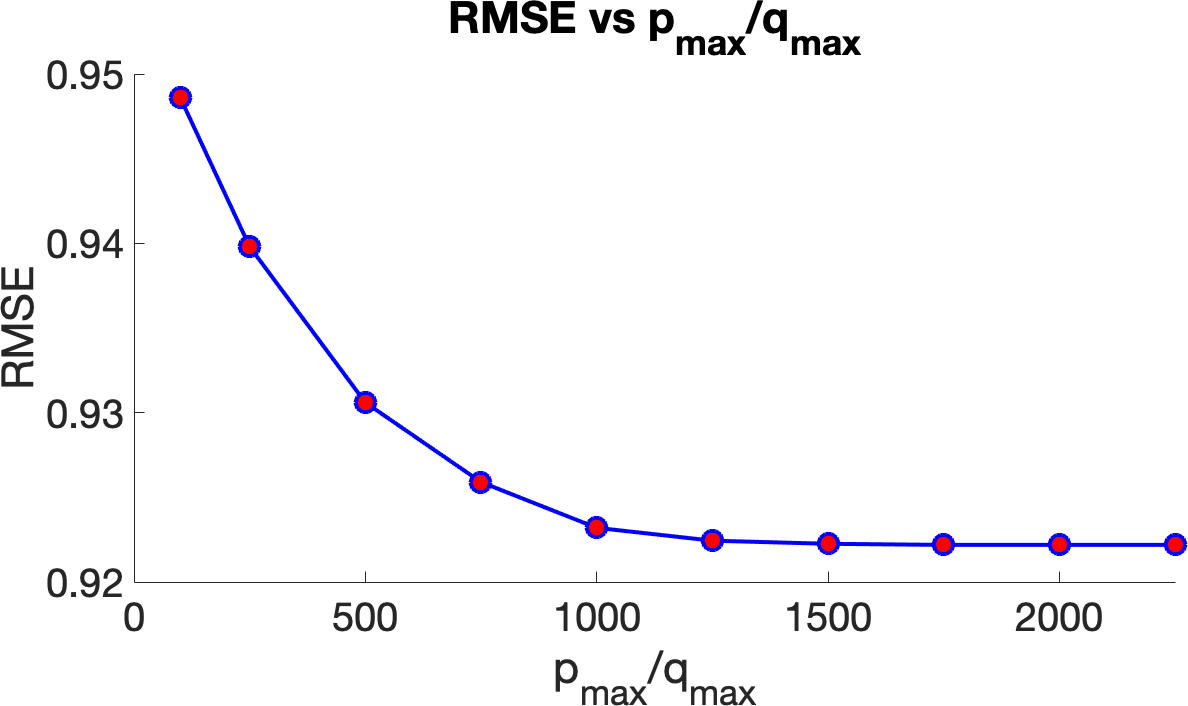}}
\caption{Effect of overparametrization: SGMC (left) and DMF (right). x-axis indicates the values of $\pmax, \qmax$, and y-axis presents the RMSE. Green X denotes the baseline from \Cref{table:results}.
}
\label{fig:ablation_DMF}
\end{figure*}

\paragraph{Ablation study.}
We study the effects of different hyper-parameters of the algorithms on the final reconstruction of the matrix. We perform an ablation study on the effects of $\diagweight, \dirweight, \pmax, \qmax$ on DMF, SGMC and SGMC-Z. The results are summarized in Figures \ref{fig:ablation_SGMC}, \ref{fig:ablation_SGMCZ}, \ref{fig:ablation_DMF}. It is interesting to note that in the case of DMF and SGMC, overparametrizing $\bb{C}, \Rpsi, \Rphi$ consistently improves the performance (see \autoref{fig:ablation_DMF}), but it only holds up to a certain point, beyond which the overparametrization does not seem to effect the reconstruction error. Notice that in the \Cref{table:params}, $\dirweight_r, \dirweight_c$ control the Dirichlet energy of rows and columns; while $\diagweight_r, \diagweight_c$ govern the weights of row/column diagonalization energy.

\begin{figure*}[!thb]
\centering
\vfill
\subfigure{\includegraphics[width=0.48\textwidth]{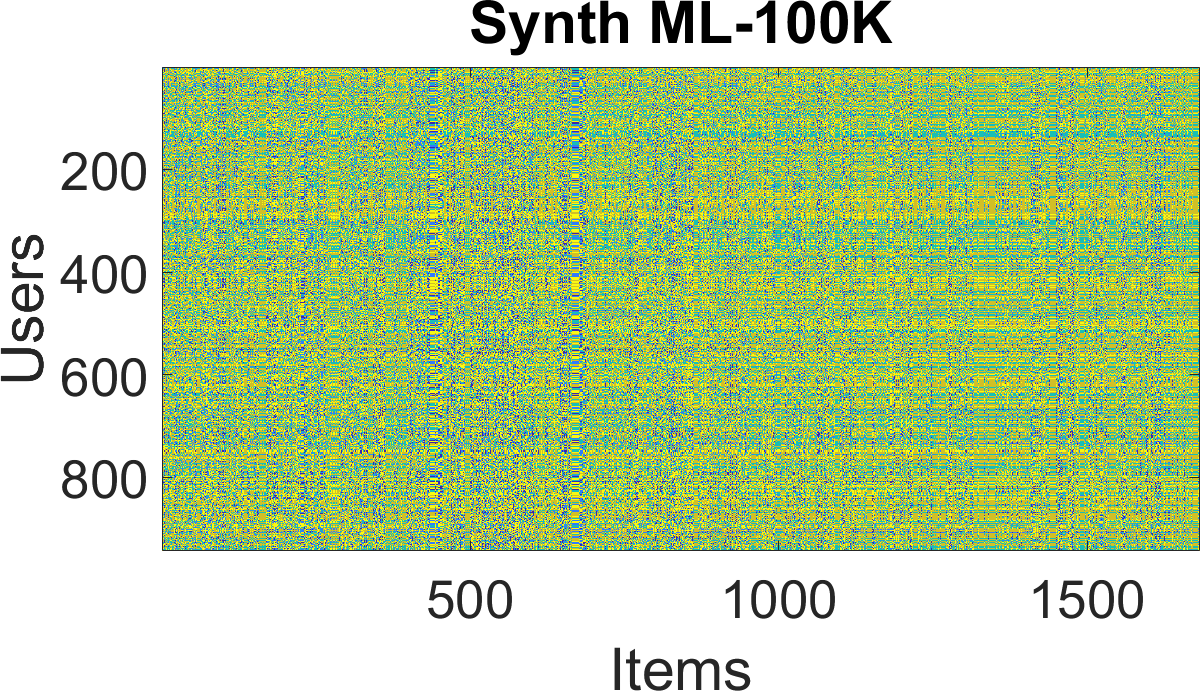}}
\subfigure{\includegraphics[width=0.48\textwidth]{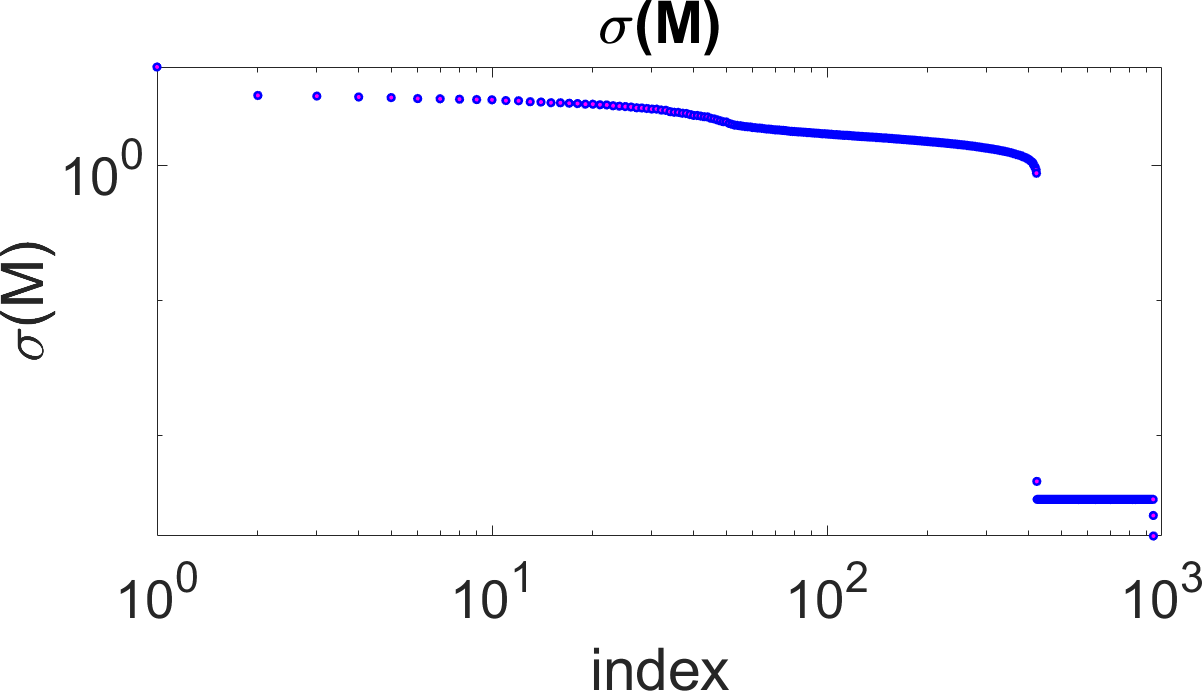}}\\
\subfigure{\includegraphics[width=0.48\textwidth]{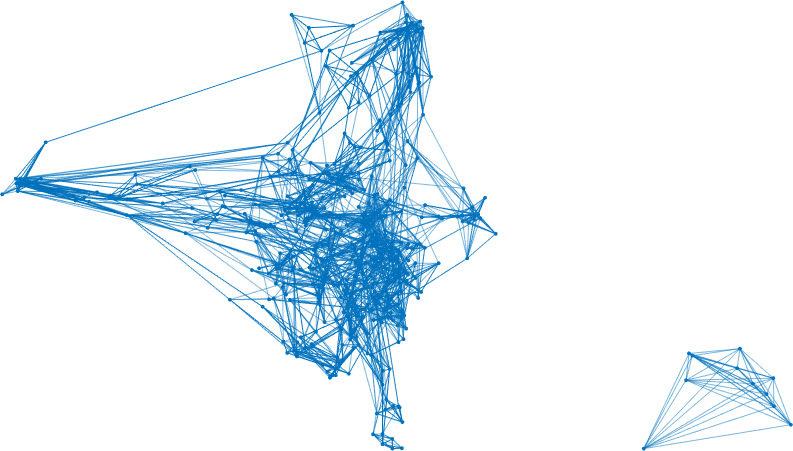}}
\subfigure{\includegraphics[width=0.48\textwidth]{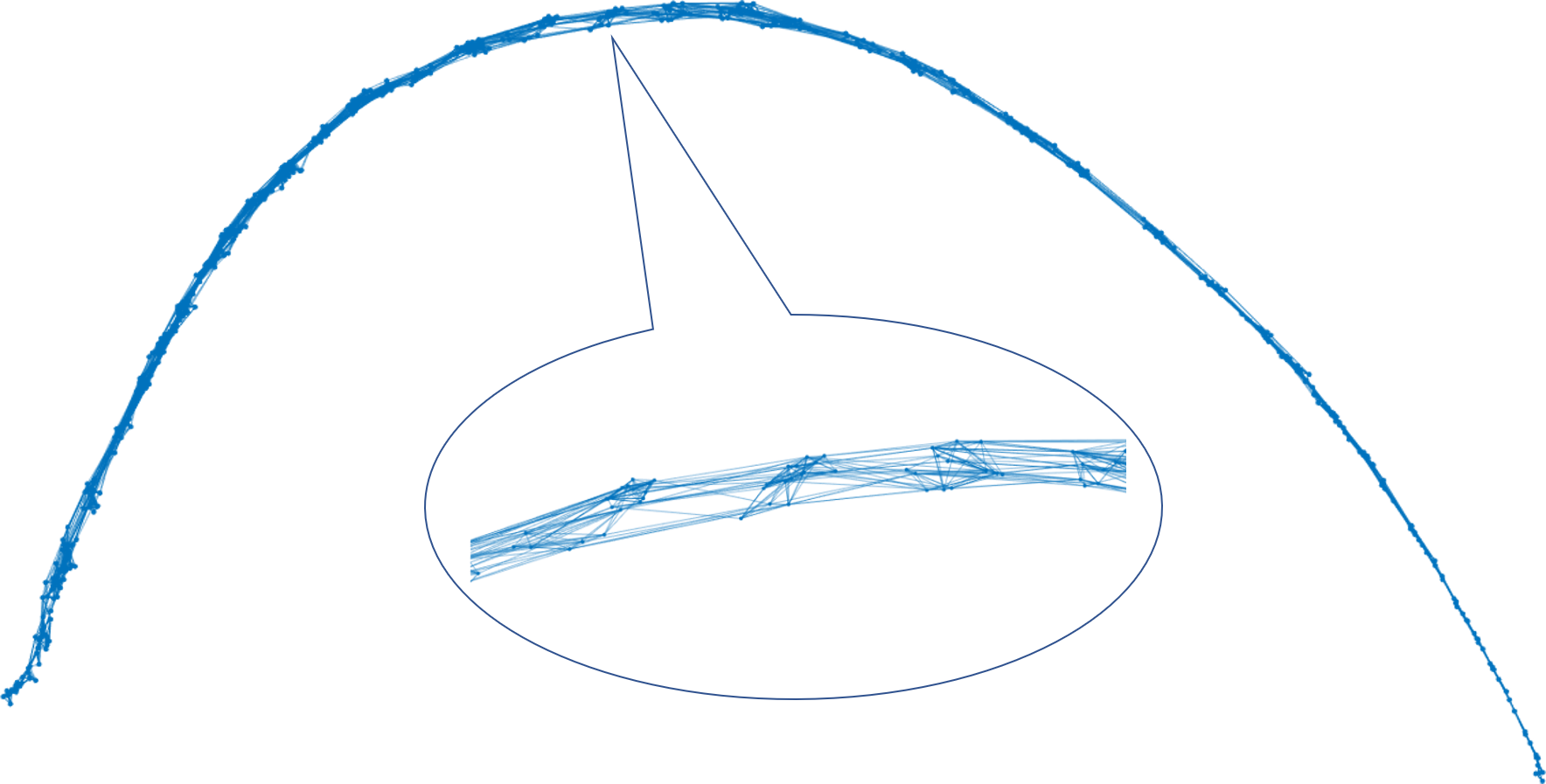}}
\caption{Synthetic Movielens-100k. Top-left: Full matrix. Top-right: singular values of the full matrix. Bottom left \& right: items \& users graph. Both graphs are constructed using 10 nearest neighbors.}
\label{fig:synth_movielens100k}
\end{figure*}

\paragraph{Synthetic MovieLens-100K.}
 While the experiments reported in \Cref{table:results} showed slight margins in favor of methods using geometry, we further experimented with a synthetic model generated from the ML-100K dataset. The purpose of this experiment is to investigate whether the results are due to the DMF model or due to the geometry as incorporated by SGMC/SGMC-Z.
 The synthetic model was generated by projecting $\bb{M}$ on the first 50 eigenvectors of $\Lrows,\Lcols$, and then matching the ratings histogram with that of the original ML-100K dataset. This nonlinear operation increased the rank of the matrix from $50$ to about $400$. See \Cref{fig:synth_movielens100k} in the \Cref{sec:appendixA} for a visualization of the full matrix, singular value distribution and the users/items graphs. The test set and training set were generated randomly and are the same size as those of the original dataset. The results reported in \Cref{tab:results_synthetic_movielens} and those on the Synthetic Netflix column in \Cref{table:results} clearly indicate that SGMC/SGMC-Z outperforms DMF, suggesting that when the geometric model is accurate it is possible to use it to improve the results. 
 
\begin{table}[!hb]
\centering
\begin{tabular}{l r}
\textbf{Model} & \textbf{ML-1M}  \\
PMF \citep{salakhutdinov2007probabilistic} & $0.883$ \\
I-RBM \citep{salakhutdinov2007restricted} & $0.854$ \\
BiasMF \citep{Koren2009matrix} & $0.845$ \\
NNMF \citep{dziugaite2015neural} & $0.843$ \\
LLORMA-Local \citep{lee2013local} & $0.833$ \\
I-AUTOREC \citep{sedhain2015autorec} & $0.831$ \\
CF-NADE \citep{zheng2016neural} & $\mathbf{0.829}$ \\
GC-MC \citep{berg2017graph} & $0.832$ \\ 
DMF \citep{arora2019implicit}, (ours) & $0.843$\\
SGMC (ours) & $0.839$\\
\end{tabular}
\caption{\small Comparison of test RMSE scores on Movielens-1M dataset. Baseline scores are taken from \citep{zheng2016neural, berg2017graph}}\label{tab:ML1M} 
\end{table}

\begin{table}[!h]
\centering
\begin{tabular}{l r}
\textbf{Model} & \textbf{Synthetic ML-100K}  \\
DMF  & $0.9147$\\
SGMC & $0.5006$\\
SGMC-Z & $\mathbf{0.4777}$ \\
\end{tabular}
\caption{\small Comparison of average RMSE of DMF, SGMC and SGMC-Z baselines calculated on 5 randomly generated Synthetic Movielens-100K datasets. }\label{tab:results_synthetic_movielens} 
\end{table}


\begin{table*}[t]
\centering
\begin{threeparttable}[!tbh]
\begin{tabular}{lrrrrrr}

\textbf{Dataset} & \textbf{Users} & \textbf{Items} & \textbf{Features} & \textbf{Ratings} & \textbf{Density} & \textbf{Rating levels}
\\[0.05em] \\[-0.8em]
Flixster  & $3,000$  & $3,000$ & Users/Items & $26,173$ & $0.0029$ & $0.5, 1, \ldots, 5$\\
Douban  & $3,000$  & $3,000$ & Users & $136,891$ & $0.0152$ & $1, 2, \ldots, 5$\\
MovieLens-100K & $943$  & $1,682$  & Users/Items & $100,000$ & $0.0630$ & $1, 2, \ldots, 5$\\
MovieLens-1M  & $6,040$  & $3,706$ & Users/Items & $1,000,209$ & $0.0447$& $1, 2, \ldots, 5$ \\
Synthetic Netflix & $150$ & $200$ & Users/Items & $4500$ & $0.15$ & $1\ldots 5$ \tnote{a}\\
Synthetic ML-100K & $943$  & $1,682$  & Users/Items & $100,000$ & $0.0630$ & $1, 2, \ldots, 5$\\
\end{tabular}

\caption{Number of users, items and ratings for Flixster, Douban, Movielens-100K, Movielens-1M, Synthetic Netflix and Synthetic Movielens-100K datasets used in our experiments and their respective rating density and rating levels.}
\begin{tablenotes}
    \item[a]{\scriptsize The ratings are not integer-valued.}
\end{tablenotes}
\label{table:datasets}
\end{threeparttable}
\end{table*}


\begin{figure*}[!hb]
\centering
\vfill
\subfigure{\includegraphics[width=0.48\textwidth]{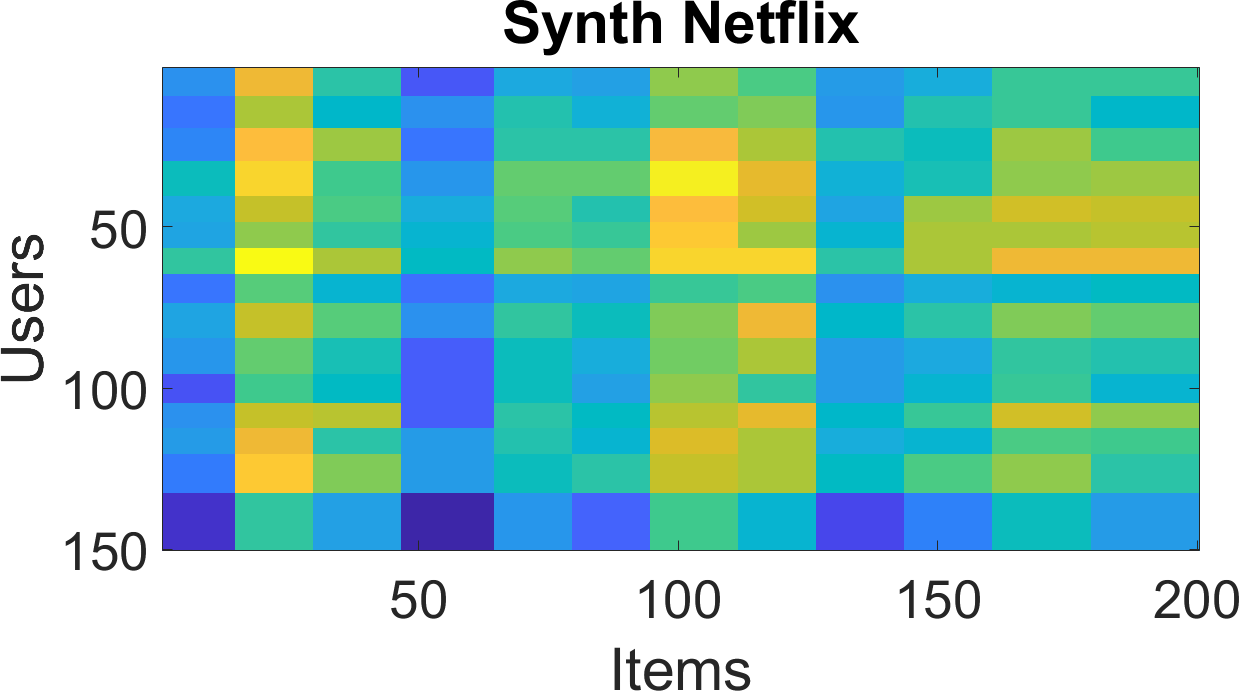}}
\subfigure{\includegraphics[width=0.48\textwidth]{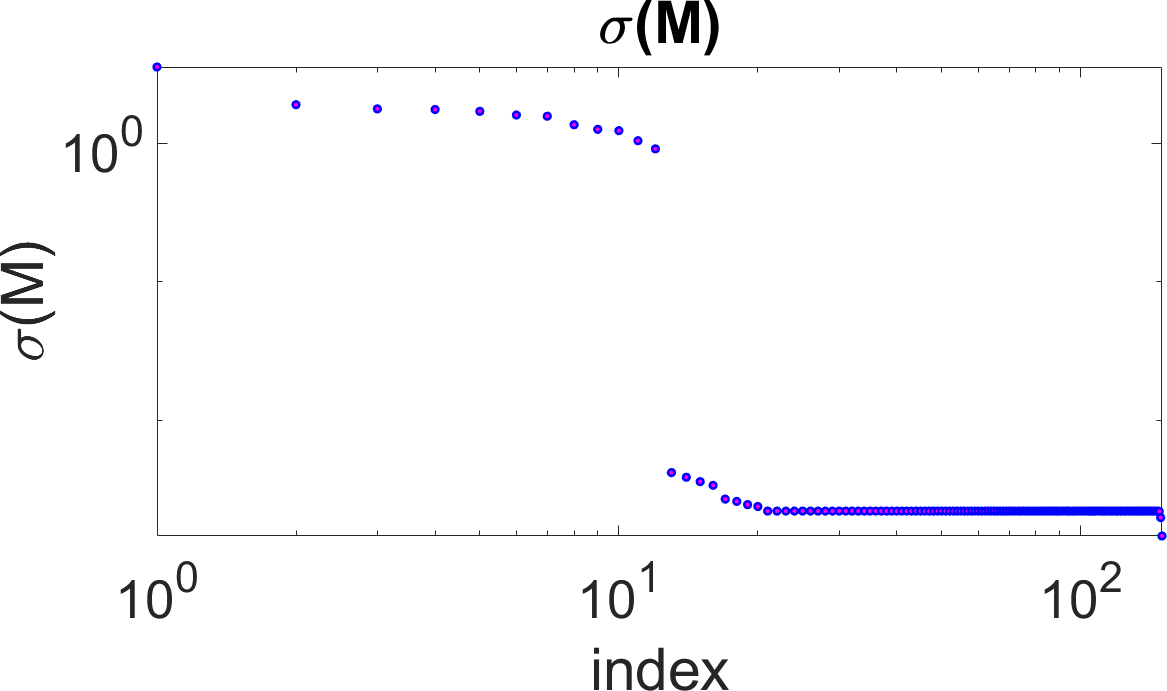}}\\
\subfigure{\includegraphics[width=0.48\textwidth]{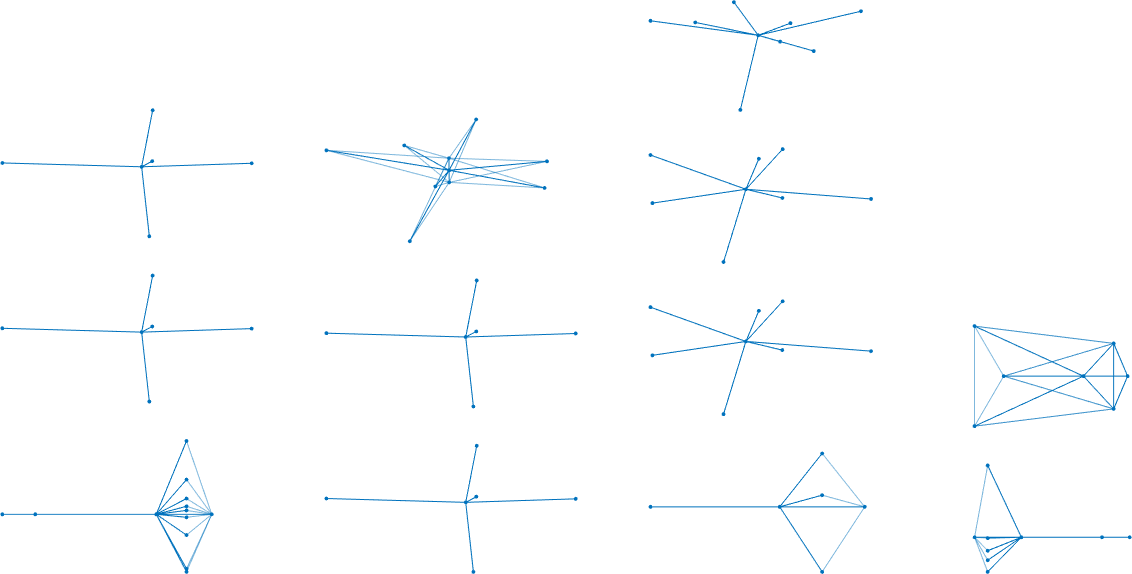}}
\subfigure{\includegraphics[width=0.48\textwidth]{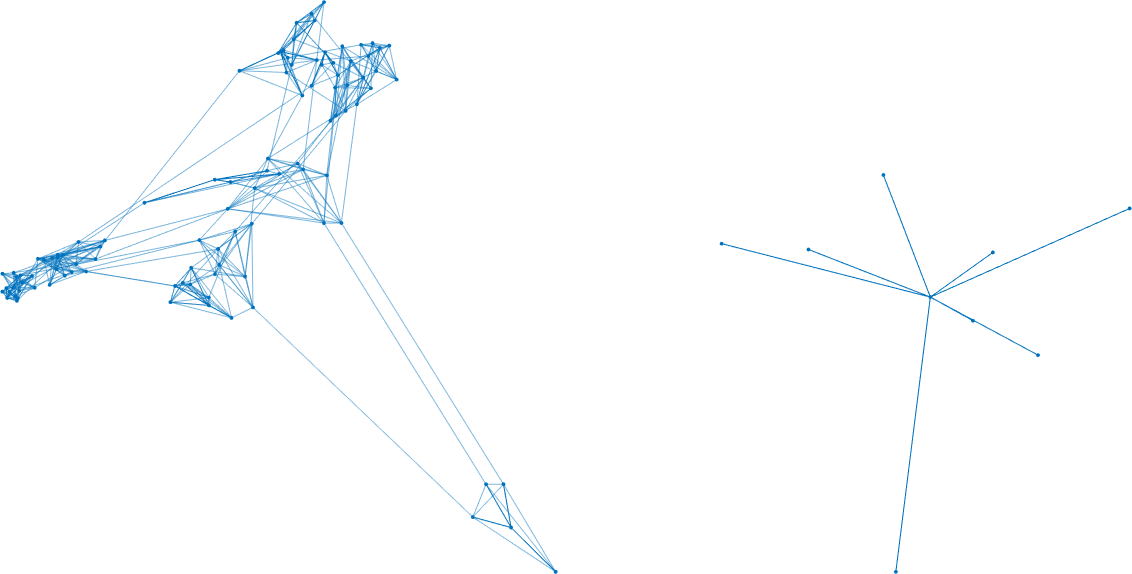}}
\caption{Synthetic Netflix. Top-left: Full matrix. Top-right: singular values of the full matrix. Bottom left \& right: items \& users graph. Taken from \citep{monti2017geometric}.}
\label{fig:synth_netflix}
\end{figure*}

\begin{table*}[!ht]
\centering
\begin{tabular}{|lrrrrrrr|}
\toprule
\textbf{Dataset} & Method & $\pmax/\qmax$ & $\pskip/\qskip$ & $\dirweight_r/\dirweight_c$ & $\diagweight_r/\diagweight_c$ &
$\substack{\text{Trainable}\\\text{variables}}$ & $\substack{\text{learning}\\\text{rate}}$
\\[0.05em] \\[-0.8em]
\toprule
 & DMF & $200/200$  & $-/-$ & $-/-$ & $-/-$ & $\Rphi,\bb{C}, \Rpsi$ & $5\times 10^{-5}$ \\
 & FM & $200/200$  & $1/1$ & $0.4/0.4$ & $-/-$ & $\bb{C}$ & $5 \times 10^{-4}$ \\
  $\Large \substack{\text{Synthetic}\\\text{Netflix}}$ & SGMC & $20/20$  & $-/-$ & $0.001/0.001$ & $0.1/-$ &  $\Rphi,\bb{C}$ & $5\times 10^{-3}$\\
  
  & SGMC-Z & $500/500$  & $3/1$ & $0.4/0.4$ & $0.1/0.1$ & $\Rphi,\bb{C}$ & $5\times 10^{-5}$ \\

\midrule
 &  DMF & $3000/3000$  & $-/-$ & $-/-$ & $-/-$ & $\Rphi,\bb{C}, \Rpsi$ & $1\times 10^{-4}$ \\
 {\text{Flixster}} & SGMC & $3000/3000$  & $-/-$ & $0.0001/0.0001$ & $0.0001/0.0001$  & $\Rphi,\bb{C}, \Rpsi$ & $1\times 10^{-4}$ \\
 
  & SGMC-Z & $200/200$  & $2/2$ & $0.0025/0.0025$ & $-/-$ & $\Rphi,\bb{C}$ &  $5\times 10^{-6}$  \\
\midrule
 {\text{Flixster}} & SGMC & $3000/3000$  & $-/-$ & $0.0001/-$ & $0.0001/-$  & $\Rphi,\bb{C}, \Rpsi$ & $5\times 10^{-5}$ \\
 (users only) & SGMC-Z & $200/200$  & $20/20$ & $0.0025/-$ & $0.001/-$ &$\Rphi,\bb{C},\Rpsi$ &  $5\times 10^{-7}$  \\
 \midrule
 & DMF & $3000/3000$  & $-/-$ & $-/-$ & $-/-$ & $\Rphi,\bb{C},\Rpsi$ &  $6\times 10^{-6}$ \\
  Douban  & SGMC & $2500/2500$  & $-/-$ & $0.001/-$ & $0.001/-$  & $\Rphi,\bb{C},\Rpsi$ &  $2\times 10^{-6}$ \\
    & SGMC-Z & $1000/1000$  & $50/1000$ & $0.011/0$ & $0.004/0$ & $\Rphi,\bb{C},\Rpsi$ &  $2\times 10^{-6}$  \\
\midrule
& DMF & $2000/2000$  & $-/-$  & $-
/-$
 & $-/-$ & $\Rphi,\bb{C},\Rpsi$ &  $5\times 10^{-5}$  \\
 ML-100K  & SGMC & $4000/4000$  & $-/-$  & $0.0003/0.0003$ & $0.0001/0.0001$ & $\Rphi,\bb{C},\Rpsi$ &  $5\times 10^{-5}$   \\
  & SGMC-Z & $3200/3200$  & $30/35$  & $0.03/0.03$ & $0.2/0.2$ & $\Rphi,\bb{C},\Rpsi$ &  $3\times 10^{-7}$   \\
 \midrule
 & DMF & $7000/7000$  & $-/-$  & $-
/-$ & $-/-$ & $\Rphi,\bb{C},\Rpsi$ &  $1\times 10^{-5}$ \\
 ML-1M & SGMC & $7000/7000$  & $-/-$  & $0.0001/0.0001$ & $-/-$ & $\Rphi,\bb{C}$ &  $8\times 10^{-5}$  \\
 
 \midrule
 & DMF & $8000/8000$  & $-/-$  & $-
/-$
 & $-/-$
 & $\Rphi,\bb{C},\Rpsi$ &  $9\times 10^{-5}$  \\
 $\Large \substack{\text{Synthetic}\\\text{ML-100K}}$ & SGMC & $600/600$  & $-/-$  & $0.001/0.001$ & $0.009/0.009$ & $\Rphi,\bb{C},\Rpsi$ &  $2\times 10^{-5}$ \\
  
  & SGMC-Z & $500/500$  & $3/1$  & $0.001/0.001$ & $0.009/0.009$ & $\Rphi,\bb{C}$ &  $5\times 10^{-6}$ \\

\bottomrule
\end{tabular}
\caption{Hyper-parameter settings for the algorithms: DMF, SGMC and SGMC-Z, reported in Tables \ref{table:results}, \ref{tab:ML1M}, \ref{tab:results_synthetic_movielens}.
\label{table:params}}
\end{table*}
 

\begin{figure*}[!ht]
\centering
\vfill
\subfigure{\includegraphics[width=0.48\textwidth]{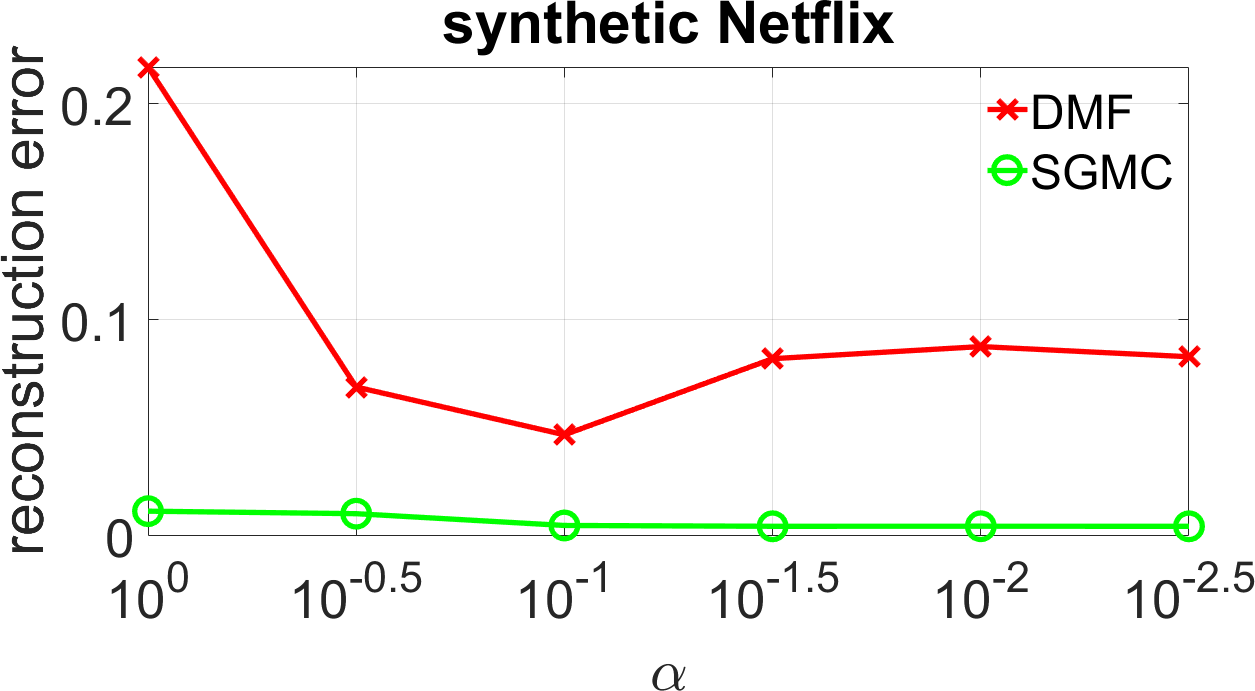}}
\subfigure{\includegraphics[width=0.48\textwidth]{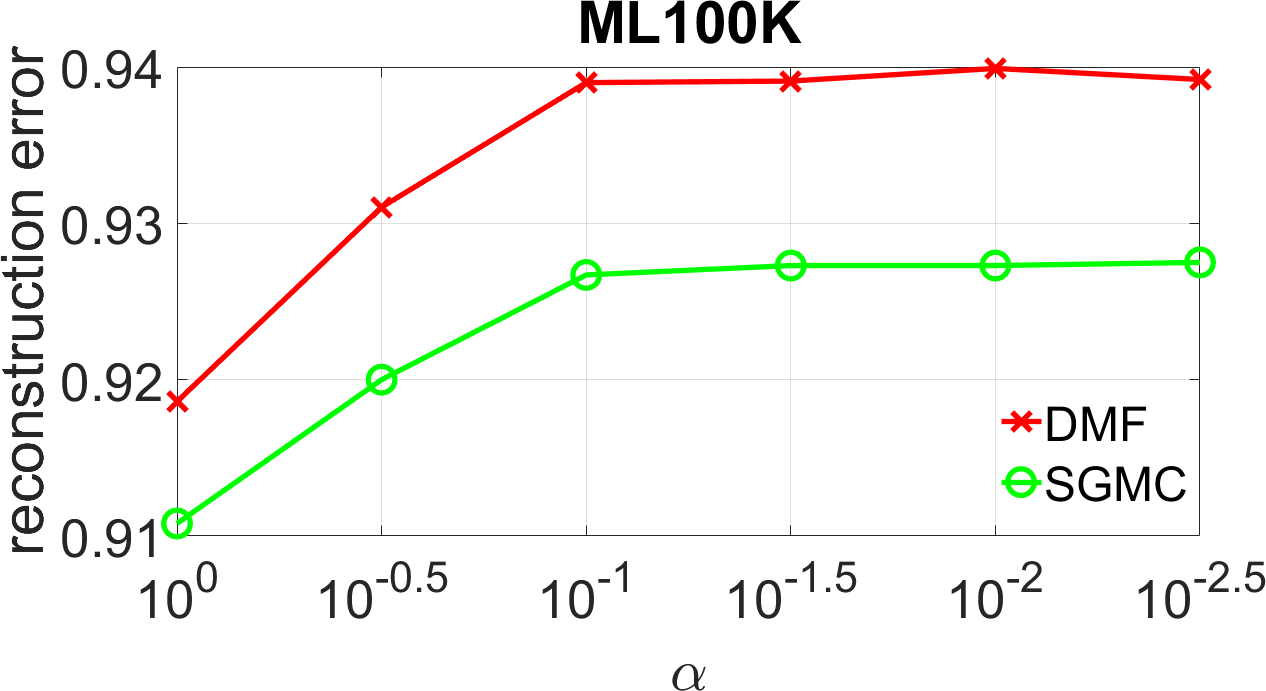}}\\
\caption{Reconstruction error (on the test set) vs. scale of initialization. For each method we initialized $\Rphi,\Rpsi$ with $\alpha\bb{I}$.
SGMC consistently outperforms DMF for any initialization. 
}
\label{fig:init}
\end{figure*}
\clearpage
\section{Drug-target interaction}
\label{sec:appendixB}
\begin{table}[!h]
    \centering
    \begin{tabular}{|l|l|l|l|l|l|}
    \hline
        Es & &  &  &  &  \\ \hline
             &      & MGRNNM & GRMF & CMF & SGMC   \\ \hline
        CVS1 & AUC  & 0.9940 $\pm$ 0.0019 & 0.9900 $\pm$ 0.0017 & 0.8443 $\pm$  0.0178 & \gr{0.9967 $\pm$ 0.0016} \\ \hline
             & AUPR & 0.9559 $\pm$ 0.0059 & 0.9295 $\pm$ 0.0081 & 0.6733 $\pm$  0.0238 & \gr{0.9729 $\pm$ 0.0033} \\ \hline
             & RMSE & 0.0441 $\pm$ 0.0008 & 0.0476 $\pm$ 0.0008 & 0.2045 $\pm$  0.0079 & \gr{0.0432 $\pm$ 0.0007} \\ \hline
        CVS2  & AUC & 0.9333 $\pm$ 0.0229 & 0.9582 $\pm$ 0.0138 & 0.9183 $\pm$  0.1068 & \gr{0.9656 $\pm$ 0.0168} \\ \hline
             & AUPR & 0.8350 $\pm$  0.0364 & 0.8553 $\pm$ 0.0284 & 0.3406 $\pm$ 0.0726 & \gr{0.8565 $\pm$ 0.0285} \\ \hline
             & RMSE & 0.0776 $\pm$ 0.0065 & 0.0827 $\pm$ 0.0066 & 0.5427 $\pm$ 0.0628 & \gr{0.0541 $\pm$ 0.0049} \\ \hline
        CVS3 & AUC  & 0.9709 $\pm$ 0.0109 & 0.9674 $\pm$ 0.0159 & 0.8525 $\pm$ 0.0188 & \gr{0.9760 $\pm$ 0.1009} \\ \hline
             & AUPR & 0.0909 $\pm$ 0.0280 & 0.9011 $\pm$ 0.0292 & 0.1958 $\pm$ 0.0636 & \gr{0.9134 $\pm$ 0.0373} \\ \hline
             & RMSE & 0.0959 $\pm$ 0.0039 & 0.0925 $\pm$ 0.0036 & 0.1842 $\pm$ 0.0370 & \gr{0.0512 $\pm$ 0.0037} \\ \hline
        GPCRs &  &  &  &  &   \\ \hline
         &  & MGRNNM & GRMF & CMF & SGMC \\ \hline
        CVS1 & AUC & 0.9770 $\pm$ 0.0068 & 0.9765 $\pm$ 0.0061 & 0.9129 $\pm$ 0.0114 & \gr{0.9831 $\pm$ 0.0065} \\ \hline
             & AUPR & 0.7995 $\pm$ 0.023 & 0.8000 $\pm$ 0.0028 & 0.7306 $\pm$ 0.0164 & \gr{0.8691 $\pm$ 0.02} \\ \hline
             & RMSE  & 0.1136 $\pm$ 0.0027 & 0.1139 $\pm$ 0.0026 &  0.7306 $\pm$ 0.0164  & \gr{0.0954 $\pm$ 0.0026} \\ \hline
        CV-B & AUC  & 0.9664 $\pm$ 0.0087 & 0.9705 $\pm$ 0.0091 & 0.9601 $\pm$ 0.0153 & \gr{0.9750 $\pm$ 0.0009} \\ \hline
             & AUPR & \gr{0.8936 $\pm$ 0.0187} & 0.8892 $\pm$ 0.0188 & 0.8754 $\pm$ 0.0364 & \red{0.8836 $\pm$ 0.0199} \\ \hline
             & RMSE & 0.1440 $\pm$ 0.0070 & 0.1476 $\pm$ 0.0065 & 0.1381 $\pm$ 0.0151 & \gr{0.1009 $\pm$ 0.0060} \\ \hline
        CVS3 & AUC  & 0.8762 $\pm$ 0.0258 & 0.9297 $\pm$ 0.0170 & 0.7843 $\pm$ 0.0701 & \gr{0.9299 $\pm$ 0.0258} \\ \hline
             & AUPR & 0.6866 $\pm$ 0.0658 & 0.7149 $\pm$ 0.0493 & 0.2256 $\pm$ 0.1021 & \gr{0.7232 $\pm$ 0.0566} \\ \hline
             & RMSE & 0.1495 $\pm$ 0.0150 & 0.1499 $\pm$ 0.0173 & 1.5743 $\pm$ 0.2302 & \gr{0.1179 $\pm$ 0.0099} \\ \hline
        ICs &  &  &  &  & \\ \hline
         &  & MGRNNM & GRMF &  CMF & SGMC  \\ \hline
        CVS1 & AUC & 0.9947 $\pm$ 0.0013 & 0.9922 $\pm$  0.0015 & 0.8745  $\pm$  0.0134 & \gr{0.9964  $\pm$  0.001} \\ \hline
         & AUPR & 0.9584  $\pm$  0.0038 & 0.9527  $\pm$  0.0043 & 0.8172  $\pm$  0.0259 & \gr{0.9784  $\pm$  0.0023} \\ \hline
         & RMSE & 0.0874  $\pm$  0.0047 & 0.0780  $\pm$  0.0021 & 0.2487  $\pm$  0.0085 & \gr{0.0710  $\pm$  0.0015} \\ \hline
        CVS2 & AUC & 0.0971  $\pm$ 0.0142 & 0.9689  $\pm$  0.0138 & 0.9229  $\pm$  0.0184 & \gr{0.9714  $\pm$  0.0156} \\ \hline
         & AUPR & 0.9026  $\pm$  0.0326 & 0.9014  $\pm$  0.0314 & 0.6426 $\pm$  0.0632 & \gr{0.9044  $\pm$ 0.0308} \\ \hline
         & RMSE & 0.1780 $\pm$  0.0118 & 0.1548  $\pm$  0.0122 & 0.4632  $\pm$  0.1578 & \gr{0.0948  $\pm$  0.0117} \\ \hline
        CVS3 & AUC & 0.9547  $\pm$  0.0188 & 0.9703  $\pm$  0.0115 & 0.7781  $\pm$  0.0344 & \gr{0.9731  $\pm$ 0.0116} \\ \hline
         & AUPR & 0.9030  $\pm$ 0.0341 & 0.9147  $\pm$ 0.0304 & 0.2198  $\pm$  0.0580 & \gr{0.9196  $\pm$  0.0264} \\ \hline
         & RMSE & \gr{0.0901  $\pm$  0.0084} & 0.1520  $\pm$  0.0045 & 0.3598  $\pm$  0.0615 & \red{0.0911 $\pm$  0.0068} \\ \hline
    \end{tabular}
    \caption{Results obtained on three drug-target interaction datasets: enzymes (Es), ion channels (ICs), G protein-coupled receptors (GPCRs). Each entry presents the mean and standard deviation across 5 runs (with different random seeds) of 10-fold cross validation. Descriptions of the evaluated baselines are reported in the text.
    Colored in green are the cases where SGMC ranks first, and in red are the cases where SGMC ranks is second or third.\label{table:DTI}}
    
\end{table}

\end{document}